\documentclass{article} 
\usepackage{iclr2024_conference,times}

\usepackage[utf8]{inputenc} 
\usepackage[T1]{fontenc}    
\usepackage{hyperref}       
\usepackage{url}            
\usepackage{booktabs}       
\usepackage{amsfonts}       
\usepackage{nicefrac}       
\usepackage{microtype}      
\usepackage{xcolor}         
\usepackage{overpic}
\usepackage{wrapfig}
\usepackage{multirow}
\usepackage{amsmath}
\usepackage{amssymb}
\usepackage{adjustbox}
\usepackage{appendix}

\newcommand{\para}[1]{\noindent{\bf #1}}

\newcommand{\R}{\mathbb{R}}
\let \bs=\mathbf
\let \set=\mathcal

\def \diag {\mathrm{diag}}

\def \path {\mathit{path}}

\def \temp {\textup{temp}}

\def \arap {\textup{arap}}
\def \acap {\textup{acap}}

\def \geo {\textup{geo}}
\def \cycle {\textup{cyc}}
\def \deform {\textup{d}}

\def \arap {\textup{arap}}

\def \exp {\textup{exp}}

\let \set = \mathcal
\let \bs = \boldsymbol

\title{GenCorres: Consistent Shape Matching via Coupled Implicit-Explicit Shape Generative Models}

\author{Haitao Yang\textsuperscript{\dag}, Xiangru Huang\textsuperscript{\ddag}, Bo Sun\textsuperscript{\dag}, Chandrajit Bajaj \textsuperscript{\dag}, Qixing Huang\textsuperscript{\dag} \\
\textsuperscript{\dag}{The University of Texas at Austin}\hspace{0.3in} \textsuperscript{\ddag}{MIT CSAIL}
}

\renewcommand{\cite}[1]{\citep{#1}}  

\iclrfinalcopy 
\begin{document}

\maketitle
\begin{abstract}
This paper introduces GenCorres, a novel unsupervised joint shape matching (JSM) approach. Our key idea is to learn a mesh generator to fit an unorganized deformable shape collection while constraining deformations between adjacent synthetic shapes to preserve geometric structures such as local rigidity and local conformality. GenCorres presents three appealing advantages over existing JSM techniques. First, GenCorres performs JSM among a synthetic shape collection whose size is much bigger than the input shapes and fully leverages the data-driven power of JSM. Second, GenCorres unifies consistent shape matching and pairwise matching (i.e., by enforcing deformation priors between adjacent synthetic shapes). Third, the generator provides a concise encoding of consistent shape correspondences. However, learning a mesh generator from an unorganized shape collection is challenging, requiring a good initialization. GenCorres addresses this issue by learning an implicit generator from the input shapes, which provides intermediate shapes between two arbitrary shapes. We introduce a novel approach for computing correspondences between adjacent implicit surfaces, which we use to regularize the implicit generator. Synthetic shapes of the implicit generator then guide initial fittings (i.e., via template-based deformation) for learning the mesh generator. 
Experimental results show that GenCorres considerably outperforms state-of-the-art JSM techniques. The synthetic shapes of GenCorres also achieve salient performance gains against state-of-the-art deformable shape generators.
\end{abstract}


\section{Introduction}

Shape matching is a long-standing problem with rich applications in texture transfer~\cite{10.1145/1186562.1015812}, compatible remeshing~\cite{10.1145/1015706.1015811}, shape morphing~\cite{Eisenberger_2021_neuromorph}, deformation transfer~\cite{sumner2004deformation}, to name just a few. It also provides foundations for analyzing and processing shape collections~\cite{KIM:2012:EC,DBLP:journals/tog/HuangWG14,Huang:2019:Tensor:Maps}. As the sizes and variations of geometric shape collections continue to grow, there are fundamental challenges in formulating and solving the shape matching problems. Pairwise approaches work for similar shape pairs and become less effective on less similar shapes. The real difficulties lie in developing suitable matching potentials (that factor out usually unknown inter-shape variations) and non-convexity in the induced non-convex optimization problems.

In contrast to pairwise matching, joint shape matching (JSM) simultaneously optimizes consistent correspondences among a shape collection~\cite{DBLP:journals/cgf/NguyenBWYG11,10.1145/2366145.2366186,KIM:2012:EC,Huang:2013:CSM,Wang:2013:IMA,DBLP:journals/tog/HuangWG14,Huang_2019_CVPR,10.1145/3386569.3392402,Huang_2020_czo}. These techniques bypass the difficulty of matching two different shapes through paths of similar shape pairs. Despite significant advances on this topic, existing approaches present three challenges. The first is to obtain a sufficiently large dataset so that each shape has neighboring shapes where shape matching succeeds. The second is that pairwise inputs are usually detached from joint matching. Third, encoding consistent dense correspondences is costly for large shape collections.

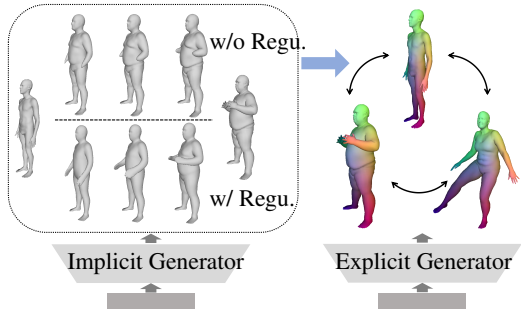
\begin{wrapfigure}{r}{0.49\textwidth} 
\centering
\begin{overpic}[width=0.5\textwidth]{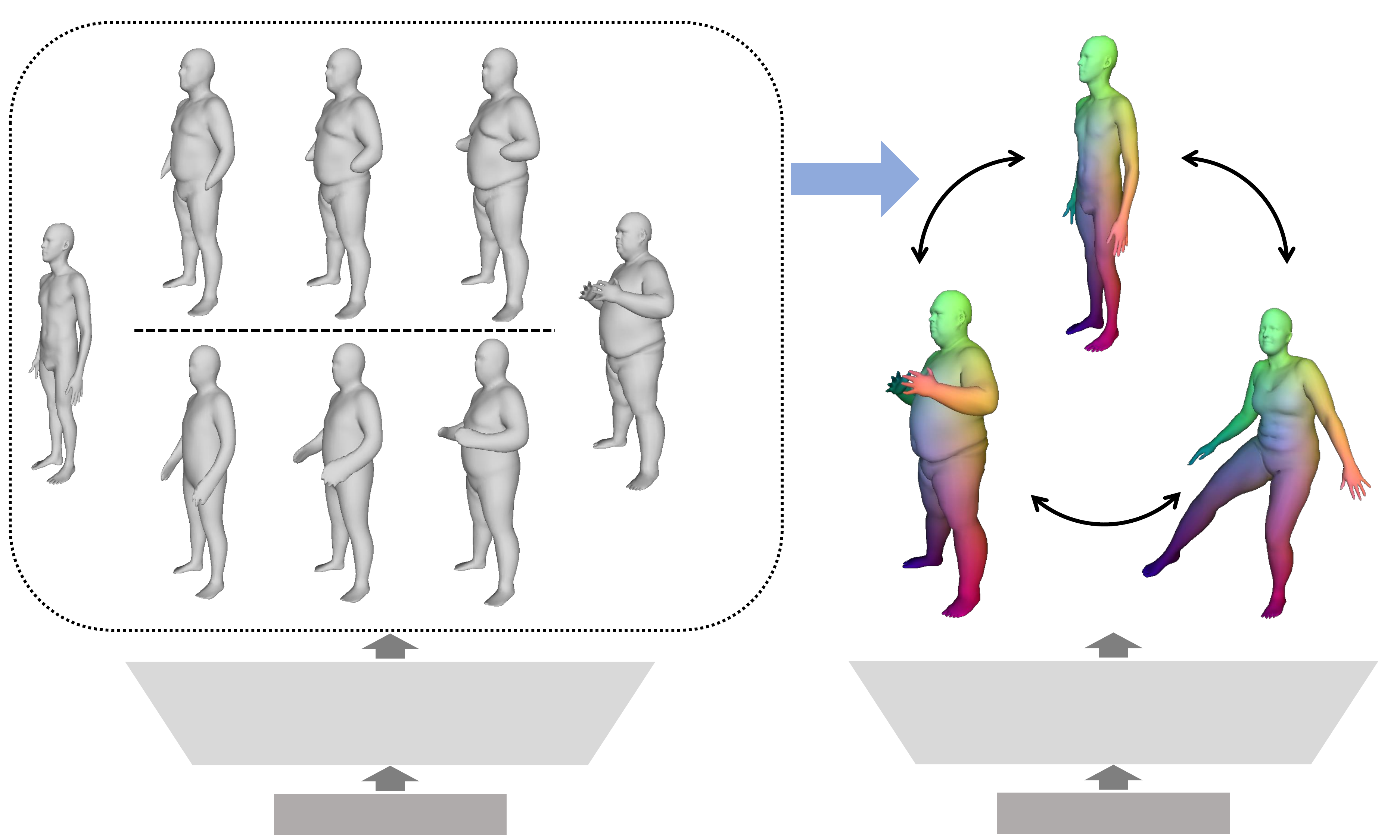}
\put(39,20){\footnotesize{w/ Regu.}}
\put(39,50){\footnotesize{w/o Regu.}}
\put(12,8){\footnotesize{Implicit Generator}}
\put(63,8){\footnotesize{Explicit Generator}}
\end{overpic}
\caption{\small{GenCorres performs consistent shape matching by learning a coupled implicit and mesh (explicit) generator to fit a shape collection without pre-defined correspondences. (Left) Interpolation between a pair of shape in the shape space. Constraining deformations between adjacent synthetic shapes with the regularization loss improves the shape space. (Right) The mesh generator provides consistent correspondences between pairs of shapes. }}
\vspace{-0.15in}
\label{Figure:GenCorres:Teaser}
\end{wrapfigure}

This paper presents \textsl{GenCorres} for solving the JSM problem. GenCorres takes motivations from recent advances in neural shape generators. Given a collection of shapes with no inter-shape correspondences, GenCorres seeks to learn a mesh generator to fit the input shapes while constraining deformations between adjacent synthetic shapes to preserve geometric structures such as local rigidity and conformality (See Figure~\ref{Figure:GenCorres:Teaser}). Interestingly, this simple framework addresses all the challenges of JSM. GenCorres performs JSM among synthetic shapes, whose size is much larger than the number of input shapes. Second, shape matching is done among neighboring shapes through the local rigidity and local conformality potentials, bypassing the difficulty of crafting a non-linear objective function between less similar shapes. In addition, GenCorres unifies pairwise matching (i.e., through deformation priors between adjacent shapes) and consistent matching (i.e., through the generator). Furthermore, the mesh generator provides an efficient encoding of shape correspondences. 

However, learning the mesh generator directly from the input shapes is challenging as it requires good initializations. Moreover, optimization procedures, e.g., that minimize the earth-mover distances between synthetic shapes and training shapes~\cite{DBLP:conf/cvpr/FanSG17,DBLP:conf/icml/AchlioptasDMG18}, can easily get trapped into local minimums. GenCorres addresses this issue by learning an implicit shape generator from the input shapes. The formulation builds on a novel approach for establishing dense correspondences between adjacent implicit surfaces defined by the shape generator. GenCorres enforces these correspondences to preserve local rigidity and conformality between pairs of adjacent shapes and satisfy the cycle consistency constraint among adjacent shape triplets. These constraints are modeled as regularization terms for learning the implicit shape generator. GenCorres then converts the learned implicit generator into an explicit mesh generator. The implicit generator offers initial consistent correspondences by guiding template-based registration. 

We have evaluated GenCorres on various deformable shape collections, including humans and animals. Experimental results show that GenCorres outperforms state-of-the-art JSM approaches and implicit and point cloud shape generators, making GenCorres a universal framework for computing JSM and learning deformable implicit shape generators. An ablation study justifies the importance of different components of GenCorres. Code is available at  \href{https://github.com/yanghtr/GenCorres}{https://github.com/yanghtr/GenCorres}.





\section{Related Work}
\label{Section:Related:Works}

We discuss relevant work under five topics, which are described below.

\noindent\textbf{Pairwise shape matching.}
Pairwise shape matching has been studied extensively in the literature~\cite{DBLP:journals/vc/Sahillioglu20,DBLP:conf/eurographics/Kaick0HC10,10.1145/2010324.1964974,Ovsjanikov:2012:FMAP,Aigerman:2014:LB,DBLP:journals/tog/MaronDKKL16,DBLP:journals/tog/MelziRRSWO19, Bednarik_2021_ICCV}. A recent line of papers establishes a learning framework under the functional map representation~\cite{DBLP:conf/iccv/LitanyRRBB17,DBLP:conf/cvpr/HalimiLRBK19,DBLP:conf/cvpr/DonatiSO20,SharmaO20-0,DBLP:journals/tog/CaoRB23}. However, existing techniques still do not work well for less similar shape pairs, where it is challenging to learn suitable matching objective functions.


NeuroMorph~\cite{Eisenberger_2021_neuromorph} combines a correspondence module with a shape interpolation module. The network is trained in an unsupervised manner. Several other methods~\cite{DBLP:journals/cgf/EisenbergerLC19,DBLP:conf/eccv/EisenbergerC20} also optimize interpolation paths to establish correspondences. While GenCorres is relevant to these approaches, GenCorres is a data-driven approach that uses an implicit generator with learned representations from all input shapes to drive pairwise matching. 


\noindent\textbf{Joint shape matching.}
The underlying principle of joint shape matching (JSM) techniques~\cite{DBLP:journals/cgf/NguyenBWYG11,10.1145/2366145.2366186,KIM:2012:EC,Huang:2013:CSM,Wang:2013:IMA,DBLP:journals/tog/HuangWG14,chen2014near,Huang_2019_CVPR,10.1145/3386569.3392402,Huang_2020_czo} is cycle-consistency.  State-of-the-art JSM techniques use the equivalence between the cycle-consistency constraint and the data matrix's low-rank property, which encodes pairwise maps in blocks (c.f.~\cite{Huang:2013:CSM}). This leads to constrained low-rank matrix recovery approaches~\cite{Huang:2013:CSM,Wang:2013:IMA,chen2014near,NIPS2017_6744,BajajGHHL18,Huang_2019_CVPR}, which possess strong theoretical guarantees.

GenCorres advances JSM in multiple ways. First, JSM's performance improves when the input collection size increases, as each shape can have more similar shapes for pairwise shape matching~\cite{DBLP:journals/cgf/NguyenBWYG11,10.1145/2366145.2366186,KIM:2012:EC,Huang:2013:CSM}. The advantage of GenCorres is that it utilizes a large collection of synthetic shapes and fully leverages the data-driven behavior of JSM. Second, in prior methods, joint matching and pairwise matching are typically decoupled. CZO~\cite{Huang_2020_czo} is an exception, yet it still requires good initializations, e.g.,~\cite{10.1145/2010324.1964974}. In contrast, GenCorres unifies pairwise matching and joint matching under a simple formulation. Cycle consistency is automatically enforced through the generator. Moreover, JSM performs pairwise matching among neighboring shapes through simple geometric regularizations. Finally, JSM still requires storing consistent matches across the input shape collection~\cite{10.1145/2366145.2366186,KIM:2012:EC,DBLP:journals/tog/HuangWG14}. GenCorres addresses this issue using a shape generator to compress consistent correspondences effectively.

\noindent\para{Generative model based correspondences.}
Generative models under explicit representations provide inter-shape correspondences, making them appealing for practical applications. However, existing methods are sub-optimal for high-fidelity correspondence computation. Most mesh-based generators~\cite{Tan0LX18,DBLP:conf/cvpr/VermaBV18,LitanyBBM18,TretschkTZGT20,RakotosaonaO20, Muralikrishnan_2022_glass} require consistent dense correspondences as input. In contrast to mesh-based generators, point-based generators~\cite{DBLP:conf/cvpr/FanSG17,DBLP:conf/icml/AchlioptasDMG18,DBLP:conf/cvpr/YangFST18,DBLP:conf/iclr/LiZZPS19,DBLP:conf/iccv/LiLFCH19} do not require inter-shape correspondences. The downside is that a point cloud is permutation-invariant. Therefore, the point indices in a point cloud do not always reflect meaningful correspondences. GenCorres addresses these limitations by performing shape matching under implicit representations using shape-preserving potentials.  


3D-CODED~\cite{DBLP:conf/eccv/GroueixFKRA18} adopts an auto-encoder to deform a template shape for shape matching. The training combines the Chamfer distance for shape alignment and regularizations on Laplacian operators and edge lengths. As the regularizations of 3D-CODED are designed for isometric deformations, and the Chamfer distance drives training, it mainly applies to minor inter-shape variations. In contrast, GenCorres applies to shape collections under large deformations. 


\noindent\textbf{Matching under implicit surfaces.}
A fundamental problem for neural implicit shape representation is defining inter-shape correspondences. The technical challenge is that there is only one constraint along the normal direction at each surface point, c.f.,~\cite{DBLP:journals/tog/StamS11}. GenCorres solves a constrained optimization problem to obtain inter-shape correspondences. A relevant formulation has been studied in~\cite{DBLP:journals/cgf/TaoSB16}. Under the implicit representation, a popular way to regularize local rigidity is the Killing vector field approach~\cite{DBLP:journals/cgf/Ben-ChenBSG10,DBLP:journals/cgf/SolomonBBG11a,DBLP:journals/cgf/TaoSB16,SlavchevaBCI17}, which is correspondence-free. In contrast, the correspondences computed by GenCorres allow us to introduce the cycle-consistency regularization.

\noindent\para{Neural implicit representations.}
Neural implicit representations have received significant interest on modeling 3D shapes, including man-made objects~\cite{DBLP:conf/cvpr/ParkFSNL19,DBLP:conf/cvpr/MeschederONNG19,DBLP:conf/cvpr/ChenZ19,DBLP:conf/cvpr/DengY021} and deformable objects~\cite{DBLP:conf/iccv/SaitoHNMLK19,DBLP:conf/cvpr/SaitoSSJ20,DBLP:conf/iccv/AlldieckXS21,DBLP:conf/cvpr/PengZXWSBZ21}. Unlike developing novel implicit network architectures, GenCorres focuses on regularization losses that enforce geometric priors for deformable objects. 

Developing regularization losses for training implicit neural networks has also been studied recently~\cite{gropp_2020_igr,DBLP:journals/corr/abs-2108-08931}. GenCorres is most relevant to~\cite{DBLP:journals/corr/abs-2108-08931}, which uses an as-killing-as-possible regularization loss to preserve global rigidity. In contrast, GenCorres focuses on maintaining local rigidity and conformality. 

GenCorres is also relevant to ARAPReg~\cite{huang2021arapreg}.
However, defining a suitable loss under the implicit representation has to address the fundamental challenge of determining inter-shape correspondences. GenCorres also enforces the cycle-consistency constraint among induced shape correspondences to enhance the implicit generator. 
\section{Problem Statement and Approach Overview}


\begin{figure}[t]
\centering
\includegraphics[width=0.95\textwidth]{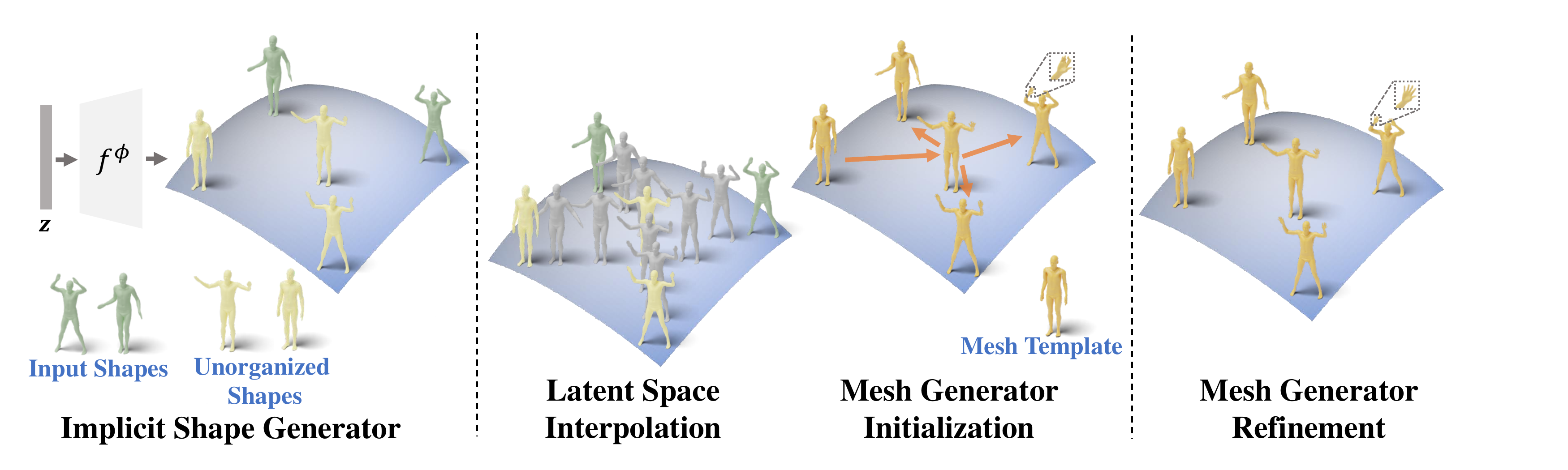}
\caption{\small{GenCorres has three stages. The first stage learns an implicit shape generator to fit the input shapes. The training loss regularizes the induced correspondences between adjacent implicit shapes of the generator. The second stage uses the implicit generator to initialize a mesh generator through latent space interpolation and template matching. The third stage then refines the mesh generator with ACAP energy. }}
\label{Figure:Approach:Overview}
\vspace{-0.2in}
\end{figure}
\noindent\para{Problem statement.}
The input to GenCorres is a shape collection $\set{S} = \{S_1,\cdots, S_{n}\}\subset \overline{\set{S}}$, where $\overline{\set{S}}$ is the underlying shape space. Each shape $S_i$ can be a raw mesh or a raw point cloud. 

GenCorres seeks to learn a mesh generator $\bs{m}^{\theta}: \set{Z}\rightarrow \overline{\set{S}}$, where $\theta$ are the network parameters, $\set{Z}:= \R^d$ is the latent space. Our goal is to align each input shape $S_i$ with the corresponding synthetic shape $\bs{m}^{\theta}(\bs{z}_i)$ where $\bs{z}_i\in \R^d$ is the latent code of $S_i$. The mesh generator then provides consistent inter-shape correspondences.  

\noindent\para{Approach overview.} As illustrated in Fig.~\ref{Figure:Approach:Overview}, GenCorres proceeds in three stages. The first two stages provide initializations for the third stage, which learns the mesh generator. Specifically, the first stage adopts variational auto-encoder (VAE) to learn an implicit generator $f^{\phi}:\R^3\times \set{Z}\rightarrow \R$ and an encoder $h^{\psi}$ from the input shapes:
\begin{align}
\min\limits_{\phi, \psi}  l_{\textup{VAE}}\big(f^{\phi}, h^{\psi}\big) + \lambda_{\geo}r_{\geo}(f^{\phi})  + \lambda_{\cycle}r_{\cycle}(f^{\phi})
\label{Eq:Implicit:Loss:Term}
\end{align}
where $\phi$ and $\psi$ are the network parameters, $l_{\textup{VAE}}$ is a VAE loss on the training shapes, $\lambda_{\geo}$ and $\lambda_{\cycle}$ are the weights of the regularization terms. $r_{\geo}(f^{\phi})$ and $r_{\cycle}(f^{\phi})$, which are key contributions of this paper, build on induced correspondences between adjacent implicit shapes defined by $f^{\phi}$. Specifically, $r_{\geo}(f^{\phi})$ enforces that the induced correspondences preserve local geometric structures. $r_{\cycle}(f^{\phi})$ enforces that the induced correspondences are cycle-consistent. In other words, $r_{\geo}(f^{\phi})$ and $r_{\cycle}(f^{\phi})$ perform pairwise matching and consistent matching, respectively.  The second stage of GenCorres fits a template mesh to all input shapes along paths of interpolated shapes provided by the implicit shape generator. The resulting correspondences are used to learn an initial mesh generator $\bs{m}^{\theta}$. The third stage of GenCorres refines the mesh generator by solving another optimization problem:
\begin{equation}
\min\limits_{\theta} d_{\exp}\big(\bs{m}^{\theta}, \overline{\set{S}}\big) + \lambda_{\deform} r_{\deform}(\bs{m}^{\theta})
\label{Eq:Explicit:Loss:Term}
\end{equation}
where $d_{\exp}\big(\bs{m}^{\theta}, \overline{\set{S}}\big)$ aligns the explicit generator with the input shape collection; $r_{\deform}(\bs{m}^{\theta})$ enforces as-conformal-as-possible deformation prior among adjacent shapes; $\lambda_{\deform}$ is the weight of $r_{\deform}(\bs{m}^{\theta})$.

\section{Stage I: Implicit Shape Generator}
\label{Section:Stage:I:Implicit}

This section introduces how to learn the implicit shape generator $f^{\phi}$. We begin with a novel approach for computing dense correspondences between adjacent implicit surfaces in Section~\ref{Subsec:Induced:Correspondence}. Based on the induced correspondences, we introduce two regularization terms $r_{\geo}(f^{\phi})$ and $r_{\cycle}(f^{\phi})$ in Section~\ref{Subsec:ARAP:ACAP} and Section~\ref{Subsec:Cycle:Consistency}, respectively. Finally, Section~\ref{Eq:Implementation:Details} elaborates on the implementation details.

\subsection{Induced Shape Correspondences}
\label{Subsec:Induced:Correspondence}

Our goal is to compute the dense correspondences between the implicit surface $f^{\phi}(\bs{x},\bs{z}) = 0$ and an adjacent implicit surface $f^{\phi}(\bs{x},\bs{z}+\epsilon\bs{v}) = 0$, where $\bs{x}\in \R^3$, $\bs{v}\in \R^d$ is a direction in the unit ball $\set{B}^d$ and $\epsilon$ is an infinitesimal value. The computation is nontrivial because of the difficulties in representing correspondences for the implicit surfaces. To this end, we first discretize $f^{\phi}(\bs{x},\bs{z})= 0$ using a mesh with $n$ vertices $\bs{g}^{\phi}(\bs{z})\in \R^{3n}$, e.g., via Marching cube~\cite{DBLP:conf/siggraph/LorensenC87}. We then formulate the corresponding vertices of $\bs{g}^{\phi}(\bs{z})$ on $f^{\phi}(\bs{x},\bs{z}+\epsilon\bs{v}) = 0$ as $\bs{g}^{\phi}(\bs{z}+\epsilon\bs{v}):=\bs{g}^{\phi}(\bs{z}) +\bs{d}^{\bs{v}}(\bs{z})\in \R^{3n}$. With this formulation, computing correspondences between two implicit surfaces is reduced to the computation of $\bs{d}^{\bs{v}}(\bs{z})$. As discussed in~\cite{DBLP:journals/tog/StamS11,DBLP:journals/cgf/TaoSB16}, for each vertex $\bs{g}_i^{\phi}(\bs{z})\in \R^3$, the implicit representation offers one constraint on its corresponding $\bs{d}_i^{\bs{v}}(\bs{z})\in \R^3$ along the normal direction:
\begin{equation}
\frac{\partial f^{\phi}}{\partial \bs{x}}(\bs{g}_i^{\phi}(\bs{z}),\bs{z})^T\bs{d}_i^{\bs{v}}(\bs{z}) +\epsilon\frac{\partial f^{\phi}}{\partial \bs{z}}(\bs{g}_i^{\phi}(\bs{z}),\bs{z})^T\bs{v} = 0.
\label{Eq:Implicit:Cons3}
\end{equation}
To introduce extra constraints on $\bs{d}^{\bs{v}}(\bs{z})$, we enforce that the displacements of the 1-ring patch at each vertex $\bs{g}_i^{\phi}(\bs{z})$ are as-rigid-as possible (ARAP)~\cite{DBLP:conf/siggraph/AlexaCL00,10.1111:j.1467-8659.2009.01380.x,huang2021arapreg} and as-conformal-as possible (ACAP)~\cite{10.1111/cgf.12451}. In the infinitesimal regime, we can approximate the latent rotation at $\bs{g}_i^{\phi}(\bs{z})$ as $I_3 + \bs{c}_i\times$. This leads to an ARAP potential on $\bs{d}^{\bs{v}}(\bs{z})$ as
\begin{align}
& \sum\limits_{i=1}^{n}\min\limits_{\bs{c}_i}\sum\limits_{j\in \set{N}_i} \|\bs{c}_i\times\big(\bs{g}_i^{\phi}(\bs{z})-\bs{g}_j^{\phi}(\bs{z})\big)-\big(\bs{d}_i^{\bs{v}}(\bs{z})-\bs{d}_j^{\bs{v}}(\bs{z}))\big\|^2  =  \ {\bs{d}^{\bs{v}}(\bs{z})}^T \overline{L}^{\arap}(\bs{g}^{\phi}(\bs{z}))\bs{d}^{\bs{v}}(\bs{z})
\label{Eq:ARAP:Implicit:Obj}
\end{align}
where the expression of $\overline{L}^{\arap}(\bs{g}^{\phi}(\bs{z}))$ is in the supp. material.


Similarly, we can parameterize the latent similarity transformation at $\bs{g}_i^{\phi}(\bs{z})$ as $(1+s_i)I_3 + \bs{c}_i\times$ and define the ACAP potential as
\begin{align}
&\sum\limits_{i=1}^{n}\min\limits_{s_i,\bs{c}_i}\sum\limits_{j\in \set{N}_i}  \|(s_iI_3+\bs{c}_i\times)\big(\bs{g}_i^{\phi}(\bs{z})-\bs{g}_j^{\phi}(\bs{z})\big)- \big(\bs{d}_i^{\bs{v}}(\bs{z})-\bs{d}_j^{\bs{v}}(\bs{z}))\big\|^2 = \ {\bs{d}^{\bs{v}}(\bs{z})}^T \overline{L}^{\acap}(\bs{g}^{\phi}(\bs{z}))\bs{d}^{\bs{v}}(\bs{z})
\label{Eq:ACAP:Implicit:Obj}
\end{align}
where the expression of $\overline{L}^{\acap}(\bs{g}^{\phi}(\bs{z}))$ is in the supp. material.

Denote $\overline{L}^{\phi}(\bs{z}) = \alpha\overline{L}^{\arap}(\bs{g}^{\phi}(\bs{z}))+\overline{L}^{\acap}(\bs{g}^{\phi}(\bs{z}))$ where $\alpha$ is a tradeoff parameter ($\alpha = 10$ in our experiments). We compute $\bs{d}^{\bs{v}}(\bs{z})$ via linearly constrained quadratic programming:
\begin{align}
\bs{d}^{\bs{v}}(\bs{z}) := \lim_{\mu\rightarrow 0} \underset{\bs{d}}{\textup{argmin}} \quad  \bs{d}^T\overline{L}^{\phi}(\bs{z}) \bs{d} + \mu\|\bs{d}\|^2 \quad
s.t. \quad C^{\phi}(\bs{z}) \bs{d} = -\epsilon F^{\phi}(\bs{z})\bs{v}
\label{Eq:Quad:Programming}
\end{align}
where $C^{\phi}(\bs{z})\bs{d} = -\epsilon F^{\phi}(\bs{z})\bs{v}$ is the matrix representation of (\ref{Eq:Implicit:Cons3}), $C^{\phi}(\bs{z})\in\R^{n\times 3n}$ is a block diagonal sparse matrix, $F^{\phi}(\bs{z})\in\R^{n\times d}$,
$\mu$ is used to avoid degenerate cases, e.g., a rotating sphere. The expressions of $C^{\phi}(\bs{z})$ and $F^{\phi}(\bs{z})$ are in the supp. material. It is easy to check that 
\begin{align}
\bs{d}^{\bs{v}}(\bs{z}) = -\epsilon G^{\phi}(\bs{z})\bs{v},\qquad 
G^{\phi}(\bs{z}) :=  {\overline{L}^{\phi}(\bs{z})}^{+}C^{\phi}(\bs{z})^T \big(C^{\phi}(\bs{z}){\overline{L}^{\phi}(\bs{z})}^{+}C^{\phi}(\bs{z})^T\big)^{+}F^{\phi}(\bs{z})   \label{Eq:Displacement:Defn}
\end{align}
where $A^{+}$ denotes the Moore–Penrose inverse of $A$.

\begin{figure}[t]
\centering
\begin{overpic}[width=0.95\textwidth]{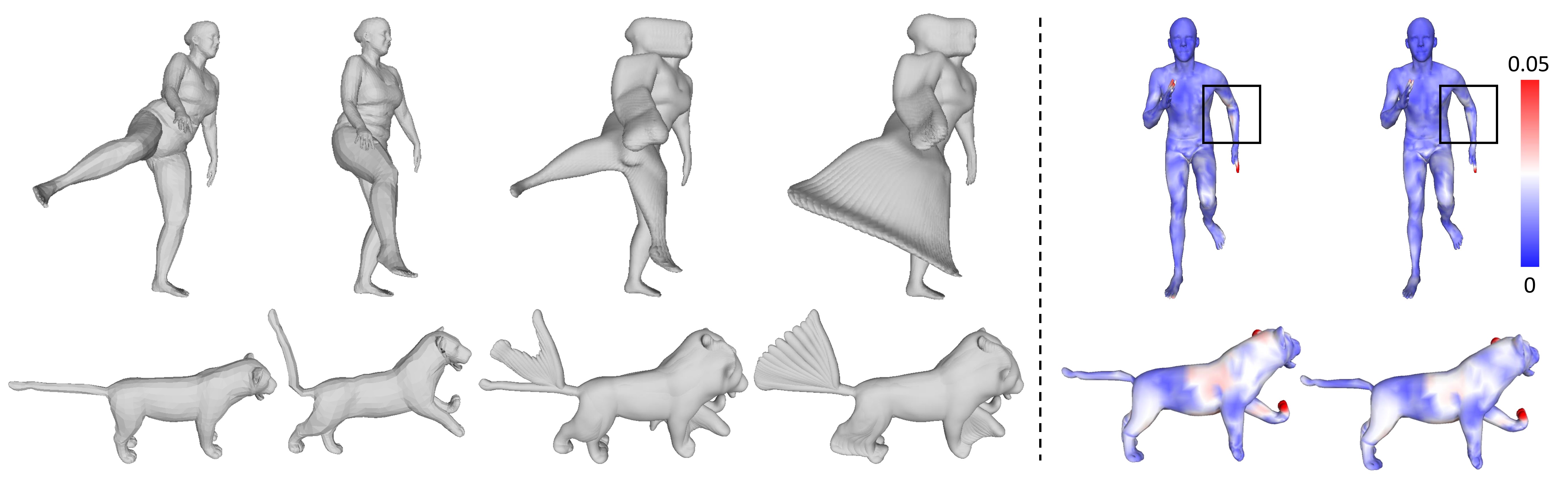}
\put(8,  0){(a)}
\put(23, 0){(b)}
\put(38, 0){(c)}
\put(55, 0){(d)}
\put(75, 0){(e)}
\put(90, 0){(f)}
\end{overpic}
\caption{\small{(Left) Effects of the geometric deformation regularization loss $r_{\geo}(f^{\phi})$. We compute 30 interpolated shapes between a source shape (a) and a target shape (b) via linear interpolation between their latent codes. All the interpolated shapes are visualized in the same coordinate system. (c) Interpolation results without $r_{\geo}(f^{\phi})$. (d) With $r_{\geo}(f^{\phi})$. (Right) Effects of the cycle-consistency regularization loss $r_{\cycle}(f^{\phi})$. We color-code errors of propagated correspondences through a path of intermediate shapes between each source-target shape pair. The error is visualized on the target mesh. (e) Without $r_{\cycle}(f^{\phi})$. (f) With $r_{\cycle}(f^{\phi}) $.}}
\label{Eq:Regularization:Effects}
\vspace{-0.1in}
\end{figure}
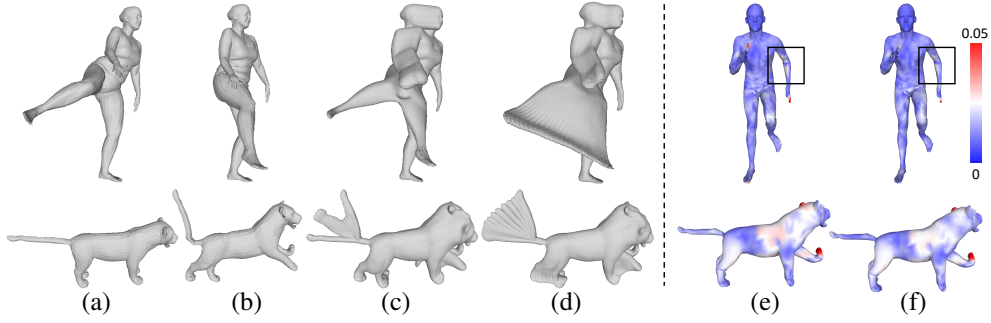



\subsection{Geometric Deformation Regularization Loss}
\label{Subsec:ARAP:ACAP}

We proceed to introduce the first regularization loss $r_{\geo}(f^{\phi})$, which penalizes local rigidity and local conformality distortions of the induced correspondences from $f^{\phi}(\bs{x},\bs{z}) = 0$ to $f^{\phi}(\bs{x},\bs{z}+\epsilon\bs{v}) = 0$:
\begin{align}
r_{\geo}(\bs{z},\bs{v}) & :=  {\bs{d}^{\bs{v}}(\bs{z})}^T \overline{L}^{\phi}(\bs{z})\bs{d}^{\bs{v}}(\bs{z}) =\epsilon^2 \bs{v}^T E^{\phi}(\bs{z}) \bs{v} 
\label{Eq:Geo:Regu:Term}  \\
E^{\phi}(\bs{z}) & :=  {F^{\phi}(\bs{z})}^T\big(C^{\phi}(\bs{z}) \overline{L}^{\phi}(\bs{z})^{+}{C^{\phi}(\bs{z})}^T\big)^{+}F^{\phi}(\bs{z})   \nonumber 
\end{align}
Integrating $\bs{v}$ over the unit ball $\set{B}^d$ in $\R^d$ ~\cite{huang2021arapreg} and omitting the constant $\epsilon^2$, we define 
\begin{align}
r_{\geo}(f^{\phi}) = \mathbb{E}_{\bs{z}\sim \set{N}_d}\int_{\bs{v}\in \set{B}_d}\bs{v}^T E^{\phi}(\bs{z}) \bs{v} d\bs{v} = \mathbb{E}_{\bs{z}\sim\set{N}_d} \frac{\text{Vol}(\mathcal{B}_d)}{d} \text{Tr}( E^{\phi}(\bs{z}) )
\label{Eq:Geo:Regu:Loss}
\end{align}
Figure~\ref{Eq:Regularization:Effects} (Left) shows that $r_{\geo}(f^{\phi})$ can improve the quality of the implicit shape generator. The interpolated shapes are smoother and more shape-preserving, leading to a better shape space .

\subsection{Cycle-Consistency Regularization Loss}
\label{Subsec:Cycle:Consistency}

The induced correspondences defined in (\ref{Eq:Displacement:Defn}) enable us to compute correspondences between two shapes by composing induced correspondences along a path of intermediate shapes. An additional regularization we can enforce is that the induced correspondences are cycle-consistent. To this end, we constrain 3-cycle consistency~\cite{Huang:2013:CSM} among three neighboring synthetic shapes $f^{\phi}(\bs{x},\bs{z}) = 0$, $f^{\phi}(\bs{x},\bs{z}+\epsilon\bs{v}) =0$, and $f^{\phi}(\bs{x},\bs{z}+\epsilon\bs{v}')= 0$, where $\bs{v}$ and $\bs{v}'$ are two different displacement vectors. Formally speaking, we model 3-cycle distortion as
\begin{align}
\bs{r}^{\bs{v},\bs{v}'}(\bs{z}) := \bs{d}^{\bs{v}}(\bs{z}) + \bs{d}^{\bs{v}'-\bs{v}}(\bs{z}+\epsilon\bs{v}) - \bs{d}^{\bs{v}'}(\bs{z}) \approx -\epsilon^2\big(\bs{v}^T\frac{\partial G^{\phi}(\bs{z})}{\partial \bs{z}}\big)(\bs{v}-\bs{v}').\label{Eq:3:Cycle:Consistency}
\end{align}
Based on (\ref{Eq:3:Cycle:Consistency}), we define the cycle-consistency regularization term as 
\begin{equation}
r_{\cycle}(f^{\phi}) = \mathbb{E}_{\bs{z}\sim \set{N}_d} \int_{\bs{v}\in \set{B}^d}\|\frac{\partial G^{\phi}(\bs{z})}{\partial \bs{z}}\|_{\set{F}}^2\cdot d\bs{v}
\label{Eq:Corres:Consistency}
\end{equation} 
where $\|\cdot\|_{\set{F}}$ is the tensor Frobienus norm. We use finite-difference to compute $r_{\cycle}(f^{\phi})$.  Specifically, we compute
$\frac{1}{\epsilon}\|G^{\phi}(\bs{z} + \epsilon_{\cycle} {\bs e}_i) - G^{\phi}(\bs{z})\|_{\set{F}}^2 $
as an approximation of $r_{\cycle}(f^{\phi})$, where ${\bs e}_i$ is a random standard basis in $\mathbb{R}^d$. In Section~\ref{Section:Experiments}, we quantitatively show that $r_{\cycle}(f^{\phi})$ further enhances the shape space.



\subsection{Implementation Details}
\label{Eq:Implementation:Details}


We use the VAE network proposed in SALD~\cite{atzmon_2021_sald}, where the encoder $h^{\psi}$ is a modified PointNet~\cite{Qi_2017_pointnet} and the decoder  $f^{\phi}$ is an 8-layer MLP. The data loss $l_{\textup{VAE}}$ is the VAE loss of SALD. We set $\lambda_{\geo} = 1e^{-3}$, $\lambda_{\cycle} = 1e^{-4}$, and $\epsilon = 1e^{-3}$. We use autograd in PyTorch~\cite{DBLP:conf/nips/PaszkeGMLBCKLGA19} to compute $F^{\phi}(\bs{z})$ and $C^{\phi}(\bs{z})$. For other derivative computations, we use finite-difference for approximation. More details are deferred to the supp. material.

\section{Stage II: Mesh Generator Initialization}

The second stage initializes the mesh generator $\bs{m}^{\theta}$ using the implicit shape generator $f^{\phi}$ obtained in the previous stage. GenCorres uses the same mesh generator as ARAPReg~\cite{huang2021arapreg}, which maps the latent code $\bs{z}$ to displacement vectors associated with vertices of a template mesh $\set{M}$. We use the learned encoder $h^{\psi}$ to find the latent code $\bs{z}_{\temp}$ of $\set{M}$. Let $\bs{z}_i = h^{\psi}(S_i)$ be the latent code of the input shape $S_i$. We generate $T$ intermediate shapes $\bs{g}^{\phi}(\bs{z}_i^{j}),1\leq j \leq T$ ($T=10$ in our experiments) by linearly interpolating $\bs{z}_{\temp}$ and $\bs{z}_i$: $\bs{z}_i^j = \bs{z}_{\temp} + j\frac{\bs{z}_i-\bs{z}_{\temp}}{T+1}$. We then apply non-rigid registration to align the template mesh $\set{M}$ with each intermediate shape $\bs{g}^{\phi}(\bs{z}_i^{j})$ in order, i.e., the alignment of one intermediate shape provides the initialization for aligning the next intermediate shape. Non-rigid alignment adopts an ARAP deformation energy, and the details are deferred to the supp. material.

After propagating the correspondences along the interpolation path in the shape space, we obtain the deformed template $\bs{m}_i^{\text{init}}$ for each input shape $S_i$. We then initialize the mesh generator $\bs{m}^{\theta}$ using the standard regression loss:
\begin{align}
\theta^\text{init} = \underset{\theta}{\textup{argmin}}\sum\limits_{i=1}^{n} \|\bs{m}_i^\text{init}-\bs{m}^{\theta}(\bs{z}_i)\|^2.
\end{align}

\section{Stage III: Mesh Generator Refinement}
\label{Section:Stage:II:Implicit:2:Explicit}


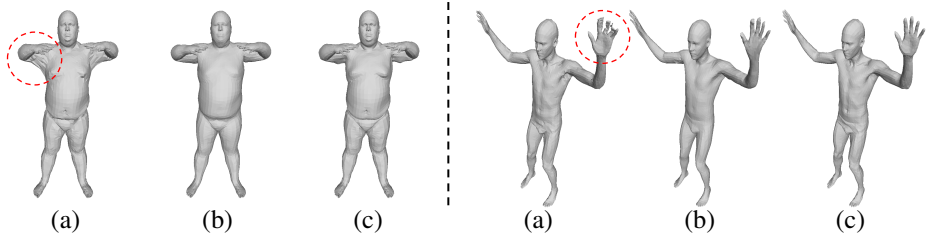
\begin{figure}
\vspace{-0.2in}
\centering
\begin{overpic}[width=0.9\textwidth]{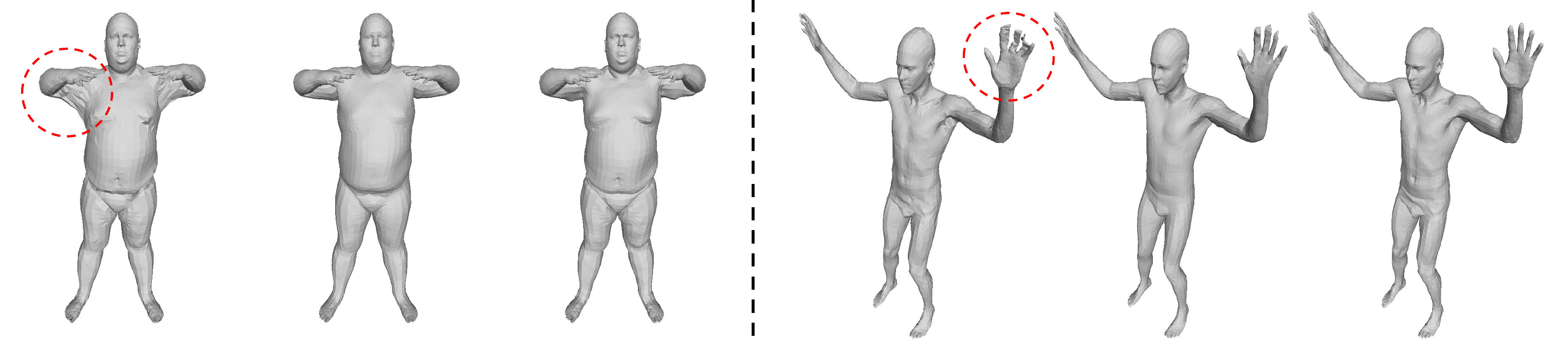}
\put(6,  -2){(a)}
\put(22, -2){(b)}
\put(38, -2){(c)}
\put(56, -2){(a)}
\put(73, -2){(b)}
\put(89, -2){(c)}
\end{overpic}
\caption{\small{The mesh generator improves the inter-shape correspondences by learning better shape generation. (a) the deformed template from stage II. (b) the shape generated by the mesh generator. (c) the input raw mesh.}}
\label{Eq:Mesh:Generator:Effects} 
\end{figure}


The third stage refines the mesh generator $\bs{m}^{\theta}(\bs{z})$ by solving (\ref{Eq:Explicit:Loss:Term}). To this end, we define the distance between the mesh generator and the input shape collection as 
\begin{equation}
d_{\exp}\big(\bs{m}^{\theta}, \overline{\set{S}}\big):= \frac{1}{n}\sum\limits_{i=1}^{n} l_\text{CD}\big( \bs{m}^{\theta}(\bs{z}_i), \set{S}_i),    
\label{Eq:Explicit:Data:Term}
\end{equation}
where $\bs{m}^{\theta}(\bs{z}_i)$ is the $i$-th generated mesh, $l_\text{CD}$ is the Chamfer loss. The loss can be optimized robustly thanks to the good initialization of the mesh generator from the first two stages. As (\ref{Eq:Explicit:Data:Term}) only constrains that vertices of the mesh generator lie on the surface, merely minimizing it does not avoid drifting. To address this issue, we define the regularization term $r_{\deform}(\bs{m}^{\theta})$ to enforce that the deformations between meshes with similar latent codes preserve geometric structures. We enforce the deformations to be ACAP, which allows the mesh generator to capture large non-rigid deformations. Based on (\ref{Eq:ACAP:Implicit:Obj}), we define 
\begin{align}
r_{\deform}(\bs{m}^{\theta}) = \mathbb{E}_{\bs{z}\sim \set{N}_d}\int_{\bs{v}\in \set{B}^d}& \bs{v}^T {\frac{\partial \bs{m}^{\theta}(\bs{z})}{\partial \bs{z}}}^T L^{\acap}(\bs{m}^{\theta}(\bs{z})){\frac{\partial \bs{m}^{\theta}(\bs{z})}{\partial \bs{z}}} \bs{v}d\bs{v}.
\label{Eq:Explicit:Regu}    
\end{align}
We then substitute (\ref{Eq:Explicit:Data:Term}) and (\ref{Eq:Explicit:Regu}) into (\ref{Eq:Explicit:Loss:Term}) to refine the mesh generator.  As shown in Figure~\ref{Eq:Mesh:Generator:Effects}, 
the mesh generator can improve the shape quality from the implicit generator. Higher shape generation quality implies better inter-shape correspondences since the mesh generator directly provides consistent correspondences. 

\section{Experimental Evaluation}
\label{Section:Experiments}

This section presents an experimental evaluation of GenCorres.
We begin with the experimental setup in Section~\ref{Subsec:Experimental:Setup}. 
Section~\ref{Subsec:Shape:Generation:Evaluation} evaluates the shape generation quality of GenCorres. We proceed to compare GenCorres with state-of-the-art JSM approaches in Section~\ref{Subsec:Joint:Shape:Matching:Evaluation}. Section~\ref{Subsec:Pairwise:Shape:Matching:Evaluation} evaluate GenCorres on FAUST~\cite{Bogo:CVPR:2014,Ren:2018:COFM}. 
Section~\ref{Subsec:Ablation}  presents an ablation study. 

\subsection{Experimental Setup}
\label{Subsec:Experimental:Setup}



\noindent\para{Datasets.}
We evaluate GenCorres on two categories of deformable shape collections, i.e., Human and Animal. The Human category considers DFAUST~\cite{dfaust:CVPR:2017} and FAUST~\cite{Bogo:CVPR:2014}. We use the registered SMPL model from the original DFAUST dataset. Since there is low variety between the adjacent shapes, we subsample 2000 meshes from the original dataset. For FAUST, we use the re-meshed version~\cite{Ren:2018:COFM}. 
Animal category has one dataset of 383 shapes~\cite{huang2021arapreg}, which is generated from SMAL~\cite{zuffi20173d}. 
Due to space constraints, we defer the details of dataset processing to the supp. material. 

\noindent\para{Evaluation protocols.} 
We evaluate the quality of shape generation by measuring the reconstruction errors of testing shapes, i.e., using the Chamfer distance between the reconstructed mesh and the original testing shape.
For correspondence evaluation, we report the mean and median geodesic errors of the predicted correspondences between involved shape pairs. 

\subsection{Evaluation on Shape Generation Quality}
\label{Subsec:Shape:Generation:Evaluation}

We compare with the state-of-the-art shape generation approaches that do not rely on pre-defined ground-truth correspondences. Those include implicit shape generators DeepSDF~\cite{DBLP:conf/cvpr/ParkFSNL19} and SALD~\cite{atzmon_2021_sald}, point-based generators, such as LGF~\cite{cai2020learning} and DPM~\cite{luo2021diffusion}. For Human category, we train the shape generator from 1000 shapes and evaluate them on the remaining 1000 shapes. For the Animal category, we use 289 shapes for training and 94 shapes for testing .

\begin{figure*}
\centering
\begin{overpic}[width=0.85\textwidth]{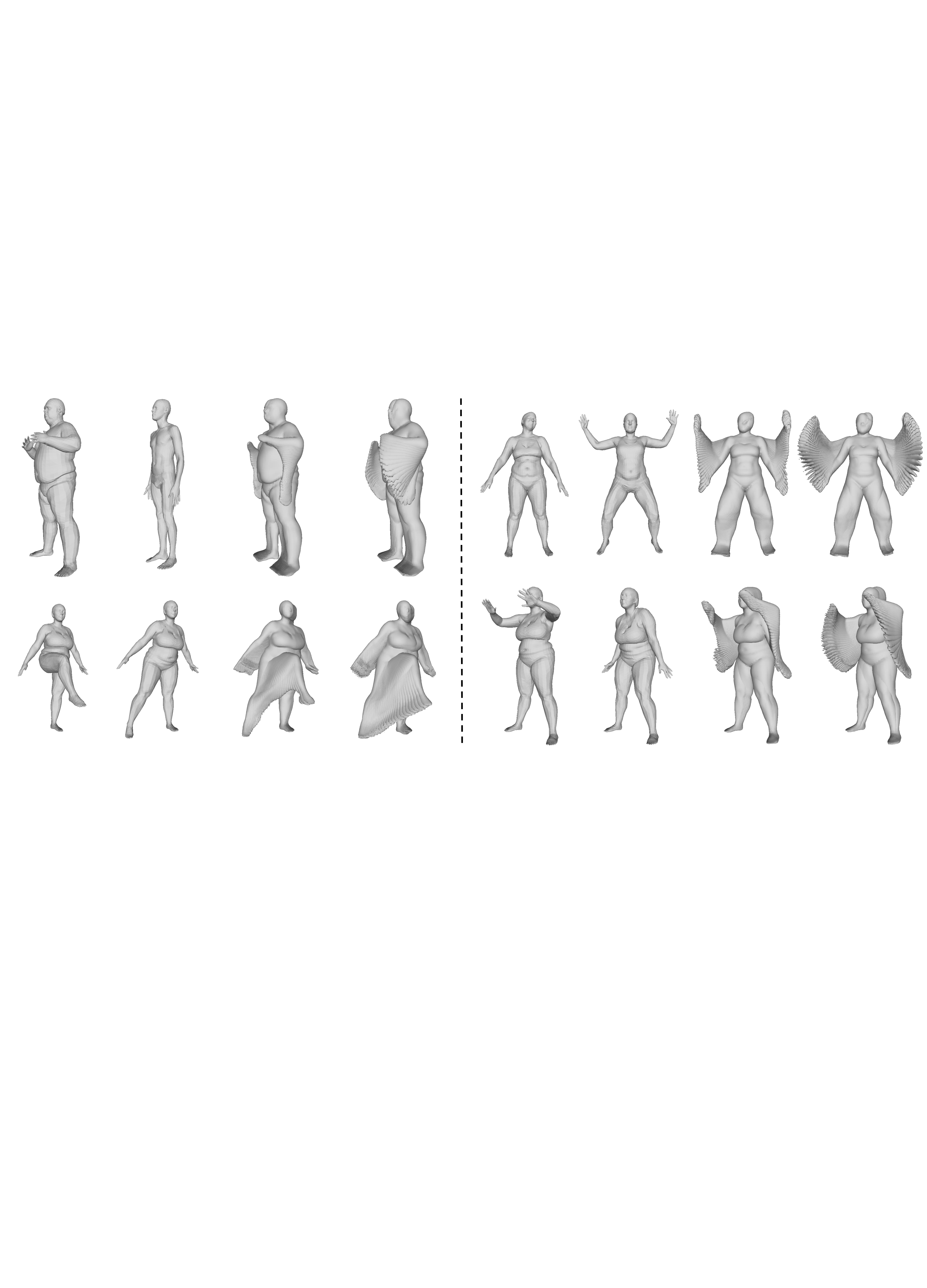}
\put(5 , -2){(a)}
\put(16, -2){(b)}
\put(29, -2){(c)}
\put(42, -2){(d)}
\put(56, -2){(a)}
\put(67, -2){(b)}
\put(80, -2){(c)}
\put(93, -2){(d)}
\end{overpic}
\caption{\small{The comparison of shape interpolation between SALD~\cite{atzmon_2021_sald} and our method on the DFAUST dataset. (a) source shape. (b) target shape. (c) interpolation results of SALD. (d) our results.}}
\label{Figure:Generation:interp_dfaust}
\end{figure*}
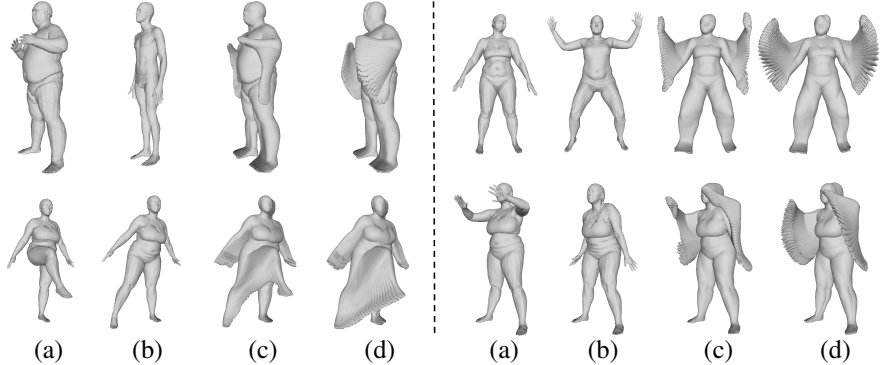

Table~\ref{Table:Generation:Quality} provides quantitative comparisons between GenCorres and baseline shape generators. Figure~\ref{Figure:Generation:interp_dfaust} shows the qualitative results. More comparisons are in the supp. material.  GenCorres is superior to all baselines in terms of both reconstruction errors and plausibility of synthetic shapes. Quantitatively, the reductions in mean/median reconstruction errors are 2.7\%/10.6\%, 2.3\%/7.0\% on DFAUST and SMAL, respectively.
Qualitatively, GenCorres provides much better interpolation results compared to SALD, especially in preserving the rigidity of arms and legs of the humans. These performance gains mainly come from the geometric deformation regularization loss employed by GenCorres. 

\setlength\tabcolsep{2pt}
\begin{table}[]
\vspace{-0.15in}
\begin{minipage}{0.48\linewidth}
\caption{\small{Evaluations of shape generation quality. For each method, we report the mean and median reconstruction errors (cm) of the testing shapes. Baselines are described in Section~\ref{Subsec:Shape:Generation:Evaluation}.\\}}
\begin{adjustbox}{width=0.99\columnwidth, center}
\centering
\small
\begin{tabular}{cl|cc|cc}
\toprule
&         & \multicolumn{2}{c|}{DFAUST} & \multicolumn{2}{c}{SMAL}  \\ \midrule 
& & mean & median & mean & median  \\ \midrule
\multirow{3}{*}{\rotatebox{0}{Point}}
& LGF                &   4.62 & 2.30  & 9.13 & 8.15 \\
& DPM               &   3.80 & 2.00  & 8.07 & 7.44 \\
\midrule
\multirow{2}{*}{\rotatebox{0}{Implicit}}
& DeepSDF  &   2.03 & 1.98  & 7.84 & 7.59 \\
& SALD              &  1.88 & 1.79  & 7.66 & 7.32 \\
\midrule
& GenCorres     &  \textbf{1.83} & \textbf{1.75}  & \textbf{6.85} & \textbf{6.81} \\ \bottomrule
\end{tabular}
\end{adjustbox}
\label{Table:Generation:Quality}
\end{minipage}\hfill
\begin{minipage}{0.48\linewidth}
\caption{\small{Evaluations of JSM on DFAUST and SMAL using geodesic errors of the predicted correspondences (in cm). Baselines are described in Section~\ref{Subsec:Joint:Shape:Matching:Evaluation}. \\}}
\begin{adjustbox}{width=0.99\columnwidth, center}
\centering
\small
\begin{tabular}{l|cc|cc}
\toprule
& \multicolumn{2}{c|}{DFAUST} & \multicolumn{2}{c}{SMAL}   \\ \midrule 
& mean  & median & mean & median \\ \midrule
CZO        & 3.71& 3.68 &   1.19 & 1.12 \\
MIM        & 3.42& 3.40 &   1.30 & 1.29\\\midrule
NeuroMorph & 2.49 & 2.47 &   1.59 & 1.43\\
\midrule
GenCorres     & \textbf{1.30} & \textbf{1.13} &   \textbf{1.02} & \textbf{0.47}\\ 
GenCorres-NoCycle & 1.41 & 1.22 & 1.11 & 0.51\\
GenCorres-NoGeoReg & 7.65 & 7.34 & 6.28 & 5.09\\
GenCorres-NoACAP & 1.62 & 1.42 & 1.07 & 0.49\\
GenCorres-Imp & 2.62 & 1.81 &   1.24 & 0.53\\
\bottomrule
\end{tabular}
\end{adjustbox}
\label{Table:Correspondence:DFAUST:Quality}
\end{minipage}
\end{table}

\subsection{Evaluation on Joint Shape Matching}
\label{Subsec:Joint:Shape:Matching:Evaluation}


Table~\ref{Table:Correspondence:DFAUST:Quality} reports statistics of GenCorres for JSM on DFAUST and SMAL. For baseline comparison, we choose consistent zoom out (CZO)~\cite{Huang_2020_czo} and multiple isometric matching (MIM)~\cite{Gao_2021_isometric}, which are two state-of-the-art JSM approaches. We evaluate the methods by computing the correspondence error between a template shape to rest of the shapes. We also report the performance of the top-performing pair-wise matching approach NeuroMorph~\cite{Eisenberger_2021_neuromorph} on these pairs. Note that NeuroMorph is originally not designed for JSM problem. Overall, GenCorres outperforms both JSM baselines by large margins. Specifically, GenCorres reduces the mean/median errors by 62.0\%/66.8\% and 14.3\%/58.0\% on DFAUST and SMAL, respectively. The performance gains come from two aspects. First, enforcing ARAP and ACAP deformations in the shape space locally is superior to applying sophisticated deformation models between pairs of shapes directly. Second, GenCorres performs map synchronization on synthetic shapes of the generator whose size is much larger than the input shape collection used by JSM baselines.

\setlength\tabcolsep{2.5pt}
\begin{table}
\caption{\small{Evaluations of pair-wise matching on FAUST dataset (in cm). Baselines are described in Section~\ref{Subsec:Pairwise:Shape:Matching:Evaluation}. Both NM and our method apply nonrigid registration as a post-processing step.}}
\begin{center}
\small
\begin{tabular}{c||ccc|ccccc|cc|c}
\toprule
 & \multicolumn{3}{c|}{Axiomatic} & \multicolumn{5}{c|}{Spectral Learning} & \multicolumn{3}{c}{Template Based}  \\ \midrule
      & BCICP & ZO  & S-Shells
      & GeoFM & AFmap  & D-Shells & ULRSSM & NM
      & 3D-CODED & IFS & Ours  
 \\ \midrule
error & 6.4 & 6.1 & 2.5
      & 1.9 & 1.9 & 1.7 & \textbf{1.6} & \textbf{1.6}
      & 2.5 & 2.6 & \textbf{1.6} \\ 
\bottomrule
\end{tabular}
\end{center}
\label{Table:Quantitative:Pairwise:Results}
\end{table}

\subsection{Evaluation on Pair-wise Shape Matching}
\label{Subsec:Pairwise:Shape:Matching:Evaluation}

Most of the shape matching approaches are evaluated on the pairwise benchmark FAUST~\cite{Bogo:CVPR:2014} with 20 testing shapes. We apply JSM on these shapes using GenCorres. However, directly applying it on 20 input shapes of FAUST does not offer satisfactory results since learning a deformable shape space from few shapes is very difficult. To show the advantage of GenCorres, we augment the input shapes with unorganized shapes from FAUST and DFAUST dataset, resulting in 1100 shapes in total. The inter-shape correspondences between two input shapes are given by the correspondences induced from the template model. 

Quantitative results are shown in Table~\ref{Table:Quantitative:Pairwise:Results}.
We mainly compare with state-of-the-art template based approaches, including 3D-CODED ~\cite{DBLP:conf/eccv/GroueixFKRA18} and IFS~\cite{DBLP:conf/eccv/SundararamanPO22}. For completeness, we also provide the results of the axiomatic methods, including  BCICP~\cite{Ren:2018:COFM}, ZO~\cite{DBLP:journals/tog/MelziRRSWO19}, and S-Shells ~\cite{Eisenberger_2020_smooth}; and the spectral learning methods, including GeoFM~\cite{DBLP:conf/cvpr/DonatiSO20}, AFmap~\cite{DBLP:journals/corr/abs-2210-06373}, ULRSSM~\cite{DBLP:journals/tog/CaoRB23}, D-Shells~\cite{Eisenberger_2020_deepshells} and NM~\cite{Eisenberger_2021_neuromorph}.
Note that template based methods do not utilize intrinsic features, thus they usually have worse performance compared to spectral learning methods, especially in the region of self-intersection.
GenCorres (Ours) outperforms all template based methods. It also achieves comparable performances to spectral learning methods. How to incorporate intrinsic features into our pipeline is left for future research.

\subsection{Ablation Study}
\label{Subsec:Ablation}

This section presents an ablation study on different components of GenCorres. As the main purpose of GenCorres is inter-shape correspondences, we focus on how the correspondence quality changes when varying different components of GenCorres (See Table~\ref{Table:Correspondence:DFAUST:Quality}). 

\noindent\para{Without the cycle-consistency regularization.} Dropping this term hurts the implicit generator. Quantitatively, the correspondence errors increase by 8.4\%/7.9\% and 8.8\%/8.5\% in mean/median on DFAUST and SMAL.  

\noindent\para{Without the geometric regularization.} The performance of GenCorres drops considerably when removing the geometric deformation regularization term. The mean/median geodesic errors increase by 488\%/549\% and 515\%/982\% on DFAUST and SMAL. This shows that even the cycle-consistency constraint is enforced on the correspondences computed from optimizing ARAP and ACAP losses, constraining that these correspondences minimize ARAP and ACAP losses is critical. 

\noindent\para{ACAP versus ARAP.}
GenCorres-NoACAP replaces ACAP regularization with the ARAP regularization.
As shown in Table~\ref{Table:Correspondence:DFAUST:Quality}, the performance of GenCorres slightly decreases. In particular, on DFAUST that exhibit large inter-shape deformations, i.e., thin versus fat and low versus tall, the performance drops are noticeable. Such performance gaps show that the ACAP regularization loss is important for modeling large non-isometric inter-shape deformations. 

\noindent\para{No explicit generator.} Finally, we drop the explicit generator and use the implicit shape generator to propagate correspondences computed along linearly interpolated intermediate shapes, i.e., GenCorres-Imp. The correspondences of the explicit generator (GenCorres) are superior to propagated correspondences of the implicit generator, i.e., 50.4\%/37.6\% and 17.7\%/11.3\% of error reductions on DFAUST and SMAL, respectively. Such improvements are expected as propagated correspondences between shapes that undergo large deformations may drift. 

\section{Conclusions, Limitations,  and Future Work}

This paper shows that learning shape generators from a collection of shapes leads to consistent inter-shape correspondences that considerably outperform state-of-the-art JSM approaches. The key novelties of GenCorres are the idea of using a mesh generator to formulate JSM and two regularization losses that enforce geometric structures are preserved and induced correspondences are cycle-consistent. We present extensive experimental results to justify the effectiveness of these two regularization terms. Besides high-quality inter-shape correspondences, GenCorres also outperforms state-of-the-art deformable shape generators trained from unorganized shape collections. 

One limitation of GenCorres is that it requires a reasonably large input shape collection to learn the shape generator and does not work with few input shapes. In this latter regime, learning pairwise matching has the advantage over GenCorres.  This issue may be partially addressed by using a more advanced implicit generator for deformable shapes, which is an area for future research.

There are ample future directions. So far, the regularization terms are based on discretizing implicit surfaces into meshes. An interesting question is how to define them without mesh discretization. Another direction is to explore regularization terms for man-made shapes, e.g., to enhance topological generalization and promote  physical stability. 

\subsubsection*{Acknowledgments}
We would like to acknowledge NSF IIS-2047677, HDR-1934932, and CCF-2019844.


\bibliography{ref,matching,haitao}

\begin{thebibliography}{84}
\providecommand{\natexlab}[1]{#1}
\providecommand{\url}[1]{\texttt{#1}}
\expandafter\ifx\csname urlstyle\endcsname\relax
  \providecommand{\doi}[1]{doi: #1}\else
  \providecommand{\doi}{doi: \begingroup \urlstyle{rm}\Url}\fi

\bibitem[Achlioptas et~al.(2018)Achlioptas, Diamanti, Mitliagkas, and Guibas]{DBLP:conf/icml/AchlioptasDMG18}
Panos Achlioptas, Olga Diamanti, Ioannis Mitliagkas, and Leonidas~J. Guibas.
\newblock Learning representations and generative models for 3d point clouds.
\newblock In \emph{{ICML}}, volume~80 of \emph{Proceedings of Machine Learning Research}, pp.\  40--49, Stockholm, Sweden, 2018. {PMLR}.

\bibitem[Aigerman et~al.(2014)Aigerman, Poranne, and Lipman]{Aigerman:2014:LB}
Noam Aigerman, Roi Poranne, and Yaron Lipman.
\newblock Lifted bijections for low distortion surface mappings.
\newblock \emph{ACM Trans. Graph.}, 33\penalty0 (4), jul 2014.
\newblock ISSN 0730-0301.
\newblock \doi{10.1145/2601097.2601158}.
\newblock URL \url{https://doi.org/10.1145/2601097.2601158}.

\bibitem[Alexa et~al.(2000)Alexa, Cohen{-}Or, and Levin]{DBLP:conf/siggraph/AlexaCL00}
Marc Alexa, Daniel Cohen{-}Or, and David Levin.
\newblock As-rigid-as-possible shape interpolation.
\newblock In Judith~R. Brown and Kurt Akeley (eds.), \emph{Proceedings of the 27th Annual Conference on Computer Graphics and Interactive Techniques, {SIGGRAPH} 2000, New Orleans, LA, USA, July 23-28, 2000}, pp.\  157--164. {ACM}, 2000.

\bibitem[Alldieck et~al.(2021)Alldieck, Xu, and Sminchisescu]{DBLP:conf/iccv/AlldieckXS21}
Thiemo Alldieck, Hongyi Xu, and Cristian Sminchisescu.
\newblock imghum: Implicit generative models of 3d human shape and articulated pose.
\newblock In \emph{{ICCV}}, pp.\  5441--5450, Montreal, Canada, 2021. {IEEE}.

\bibitem[Atzmon \& Lipman(2020)Atzmon and Lipman]{Atzmon_2020_sal}
Matan Atzmon and Yaron Lipman.
\newblock Sal: Sign agnostic learning of shapes from raw data.
\newblock In \emph{IEEE/CVF Conference on Computer Vision and Pattern Recognition (CVPR)}, June 2020.

\bibitem[Atzmon \& Lipman(2021)Atzmon and Lipman]{atzmon_2021_sald}
Matan Atzmon and Yaron Lipman.
\newblock {SALD:} sign agnostic learning with derivatives.
\newblock In \emph{9th International Conference on Learning Representations, {ICLR} 2021}, 2021.

\bibitem[Atzmon et~al.(2021)Atzmon, Novotn{\'{y}}, Vedaldi, and Lipman]{DBLP:journals/corr/abs-2108-08931}
Matan Atzmon, David Novotn{\'{y}}, Andrea Vedaldi, and Yaron Lipman.
\newblock Augmenting implicit neural shape representations with explicit deformation fields, 2021.

\bibitem[Bajaj et~al.(2018)Bajaj, Gao, He, Huang, and Liang]{BajajGHHL18}
Chandrajit Bajaj, Tingran Gao, Zihang He, Qixing Huang, and Zhenxiao Liang.
\newblock {SMAC:} simultaneous mapping and clustering using spectral decompositions.
\newblock In \emph{Proceedings of the 35th International Conference on Machine Learning, {ICML} 2018}, pp.\  334--343, Stockholmsm{\"{a}}ssan, Stockholm, Sweden, 2018. {PMLR}.
\newblock URL \url{http://proceedings.mlr.press/v80/bajaj18a.html}.

\bibitem[Bednarik et~al.(2021)Bednarik, Kim, Chaudhuri, Parashar, Salzmann, Fua, and Aigerman]{Bednarik_2021_ICCV}
Jan Bednarik, Vladimir~G. Kim, Siddhartha Chaudhuri, Shaifali Parashar, Mathieu Salzmann, Pascal Fua, and Noam Aigerman.
\newblock Temporally-coherent surface reconstruction via metric-consistent atlases.
\newblock In \emph{Proceedings of the IEEE/CVF International Conference on Computer Vision (ICCV)}, pp.\  10458--10467, October 2021.

\bibitem[Ben{-}Chen et~al.(2010)Ben{-}Chen, Butscher, Solomon, and Guibas]{DBLP:journals/cgf/Ben-ChenBSG10}
Mirela Ben{-}Chen, Adrian Butscher, Justin Solomon, and Leonidas~J. Guibas.
\newblock On discrete killing vector fields and patterns on surfaces.
\newblock \emph{Comput. Graph. Forum}, 29\penalty0 (5):\penalty0 1701--1711, 2010.

\bibitem[Bogo et~al.(2014)Bogo, Romero, Loper, and Black]{Bogo:CVPR:2014}
Federica Bogo, Javier Romero, Matthew Loper, and Michael~J. Black.
\newblock {FAUST}: Dataset and evaluation for {3D} mesh registration.
\newblock In \emph{Proceedings IEEE Conf. on Computer Vision and Pattern Recognition (CVPR)}, Piscataway, NJ, USA, June 2014. IEEE.

\bibitem[Bogo et~al.(2017)Bogo, Romero, Pons{-}Moll, and Black]{dfaust:CVPR:2017}
Federica Bogo, Javier Romero, Gerard Pons{-}Moll, and Michael~J. Black.
\newblock Dynamic {FAUST:} registering human bodies in motion.
\newblock In \emph{{CVPR}}, pp.\  5573--5582, Washington, DC, USA, 2017. {IEEE} Computer Society.

\bibitem[Cai et~al.(2020)Cai, Yang, Averbuch-Elor, Hao, Belongie, Snavely, and Hariharan]{cai2020learning}
Ruojin Cai, Guandao Yang, Hadar Averbuch-Elor, Zekun Hao, Serge Belongie, Noah Snavely, and Bharath Hariharan.
\newblock Learning gradient fields for shape generation.
\newblock In \emph{Computer Vision--ECCV 2020: 16th European Conference, Glasgow, UK, August 23--28, 2020, Proceedings, Part III 16}, pp.\  364--381. Springer, 2020.

\bibitem[Cao et~al.(2023)Cao, Roetzer, and Bernard]{DBLP:journals/tog/CaoRB23}
Dongliang Cao, Paul Roetzer, and Florian Bernard.
\newblock Unsupervised learning of robust spectral shape matching.
\newblock \emph{{ACM} Trans. Graph.}, 42\penalty0 (4):\penalty0 132:1--132:15, 2023.
\newblock \doi{10.1145/3592107}.
\newblock URL \url{https://doi.org/10.1145/3592107}.

\bibitem[Chen et~al.(2014)Chen, Guibas, and Huang]{chen2014near}
Yuxin Chen, Leonidas~J. Guibas, and Qi{-}Xing Huang.
\newblock Near-optimal joint object matching via convex relaxation.
\newblock In \emph{{ICML}}, volume~32 of \emph{{JMLR} Workshop and Conference Proceedings}, pp.\  100--108, Beijing, China, 2014. JMLR.org.

\bibitem[Chen \& Zhang(2019)Chen and Zhang]{DBLP:conf/cvpr/ChenZ19}
Zhiqin Chen and Hao Zhang.
\newblock Learning implicit fields for generative shape modeling.
\newblock In \emph{{IEEE} Conference on Computer Vision and Pattern Recognition, {CVPR} 2019, Long Beach, CA, USA, June 16-20, 2019}, pp.\  5939--5948, Long Beach, CA, USA, 2019. Computer Vision Foundation / {IEEE}.
\newblock \doi{10.1109/CVPR.2019.00609}.
\newblock URL \url{http://openaccess.thecvf.com/content\_CVPR\_2019/html/Chen\_Learning\_Implicit\_Fields\_for\_Generative\_Shape\_Modeling\_CVPR\_2019\_paper.html}.

\bibitem[Deng et~al.(2021)Deng, Yang, and Tong]{DBLP:conf/cvpr/DengY021}
Yu~Deng, Jiaolong Yang, and Xin Tong.
\newblock Deformed implicit field: Modeling 3d shapes with learned dense correspondence.
\newblock In \emph{{CVPR}}, pp.\  10286--10296, Virtual, 2021. Computer Vision Foundation / {IEEE}.

\bibitem[Donati et~al.(2020)Donati, Sharma, and Ovsjanikov]{DBLP:conf/cvpr/DonatiSO20}
Nicolas Donati, Abhishek Sharma, and Maks Ovsjanikov.
\newblock Deep geometric functional maps: Robust feature learning for shape correspondence.
\newblock In \emph{{CVPR}}, pp.\  8589--8598, Washington, DC, NY, 2020. Computer Vision Foundation / {IEEE}.

\bibitem[Eisenberger \& Cremers(2020)Eisenberger and Cremers]{DBLP:conf/eccv/EisenbergerC20}
Marvin Eisenberger and Daniel Cremers.
\newblock Hamiltonian dynamics for real-world shape interpolation.
\newblock In Andrea Vedaldi, Horst Bischof, Thomas Brox, and Jan{-}Michael Frahm (eds.), \emph{Computer Vision - {ECCV} 2020 - 16th European Conference, Glasgow, UK, August 23-28, 2020, Proceedings, Part {IV}}, volume 12349 of \emph{Lecture Notes in Computer Science}, pp.\  179--196, Virtual, 2020. Springer.
\newblock \doi{10.1007/978-3-030-58548-8\_11}.
\newblock URL \url{https://doi.org/10.1007/978-3-030-58548-8\_11}.

\bibitem[Eisenberger et~al.(2019)Eisenberger, L{\"{a}}hner, and Cremers]{DBLP:journals/cgf/EisenbergerLC19}
Marvin Eisenberger, Zorah L{\"{a}}hner, and Daniel Cremers.
\newblock Divergence-free shape correspondence by deformation.
\newblock \emph{Comput. Graph. Forum}, 38\penalty0 (5):\penalty0 1--12, 2019.

\bibitem[Eisenberger et~al.(2020{\natexlab{a}})Eisenberger, Lahner, and Cremers]{Eisenberger_2020_smooth}
Marvin Eisenberger, Zorah Lahner, and Daniel Cremers.
\newblock Smooth shells: Multi-scale shape registration with functional maps.
\newblock In \emph{IEEE/CVF Conference on Computer Vision and Pattern Recognition (CVPR)}, June 2020{\natexlab{a}}.

\bibitem[Eisenberger et~al.(2020{\natexlab{b}})Eisenberger, Toker, Leal{-}Taix{\'{e}}, and Cremers]{Eisenberger_2020_deepshells}
Marvin Eisenberger, Aysim Toker, Laura Leal{-}Taix{\'{e}}, and Daniel Cremers.
\newblock Deep shells: Unsupervised shape correspondence with optimal transport.
\newblock In Hugo Larochelle, Marc'Aurelio Ranzato, Raia Hadsell, Maria{-}Florina Balcan, and Hsuan{-}Tien Lin (eds.), \emph{Advances in Neural Information Processing Systems 33: Annual Conference on Neural Information Processing Systems 2020, NeurIPS 2020, December 6-12, 2020}, pp.\  10491--10502, Virtual, 2020{\natexlab{b}}. Curran Associates, Inc.
\newblock URL \url{https://proceedings.neurips.cc/paper/2020/hash/769c3bce651ce5feaa01ce3b75986420-Abstract.html}.

\bibitem[Eisenberger et~al.(2021)Eisenberger, Novotny, Kerchenbaum, Labatut, Neverova, Cremers, and Vedaldi]{Eisenberger_2021_neuromorph}
Marvin Eisenberger, David Novotny, Gael Kerchenbaum, Patrick Labatut, Natalia Neverova, Daniel Cremers, and Andrea Vedaldi.
\newblock Neuromorph: Unsupervised shape interpolation and correspondence in one go.
\newblock In \emph{Proceedings of the IEEE/CVF Conference on Computer Vision and Pattern Recognition (CVPR)}, pp.\  7473--7483, Virtual, June 2021. Computer Vision Foundation / {IEEE}.

\bibitem[Fan et~al.(2017)Fan, Su, and Guibas]{DBLP:conf/cvpr/FanSG17}
Haoqiang Fan, Hao Su, and Leonidas~J. Guibas.
\newblock A point set generation network for 3d object reconstruction from a single image.
\newblock In \emph{{CVPR}}, pp.\  2463--2471, Honolulu, Hawaii, 2017. {IEEE} Computer Society.

\bibitem[Gao et~al.(2021)Gao, Lahner, Thunberg, Cremers, and Bernard]{Gao_2021_isometric}
Maolin Gao, Zorah Lahner, Johan Thunberg, Daniel Cremers, and Florian Bernard.
\newblock Isometric multi-shape matching.
\newblock In \emph{Proceedings of the IEEE/CVF Conference on Computer Vision and Pattern Recognition (CVPR)}, pp.\  14183--14193, Virtual, June 2021. Computer Vision Foundation / {IEEE}.

\bibitem[Garland \& Heckbert(1997)Garland and Heckbert]{GarlandH_1997_sim}
Michael Garland and Paul~S. Heckbert.
\newblock Surface simplification using quadric error metrics.
\newblock In \emph{Proceedings of the 24th Annual Conference on Computer Graphics and Interactive Techniques, {SIGGRAPH} 1997, Los Angeles, CA, USA, August 3-8, 1997}, pp.\  209--216. {ACM}, 1997.

\bibitem[Gropp et~al.(2020)Gropp, Yariv, Haim, Atzmon, and Lipman]{gropp_2020_igr}
Amos Gropp, Lior Yariv, Niv Haim, Matan Atzmon, and Yaron Lipman.
\newblock Implicit geometric regularization for learning shapes.
\newblock In \emph{Proceedings of the 37th International Conference on Machine Learning, {ICML} 2020, 13-18 July 2020, Virtual Event}, volume 119 of \emph{Proceedings of Machine Learning Research}, pp.\  3789--3799. {PMLR}, 2020.
\newblock URL \url{http://proceedings.mlr.press/v119/gropp20a.html}.

\bibitem[Groueix et~al.(2018)Groueix, Fisher, Kim, Russell, and Aubry]{DBLP:conf/eccv/GroueixFKRA18}
Thibault Groueix, Matthew Fisher, Vladimir~G. Kim, Bryan~C. Russell, and Mathieu Aubry.
\newblock 3d-coded: 3d correspondences by deep deformation.
\newblock In \emph{{ECCV} {(2)}}, volume 11206 of \emph{Lecture Notes in Computer Science}, pp.\  235--251, Virtual, 2018. Springer.

\bibitem[Halimi et~al.(2019)Halimi, Litany, Rodol{\`{a}}, Bronstein, and Kimmel]{DBLP:conf/cvpr/HalimiLRBK19}
Oshri Halimi, Or~Litany, Emanuele Rodol{\`{a}}, Alexander~M. Bronstein, and Ron Kimmel.
\newblock Unsupervised learning of dense shape correspondence.
\newblock In \emph{{CVPR}}, pp.\  4370--4379, Long Beach, CA, 2019. Computer Vision Foundation / {IEEE}.

\bibitem[Huang \& Guibas(2013)Huang and Guibas]{Huang:2013:CSM}
Qi{-}Xing Huang and Leonidas~J. Guibas.
\newblock Consistent shape maps via semidefinite programming.
\newblock \emph{Comput. Graph. Forum}, 32\penalty0 (5):\penalty0 177--186, 2013.
\newblock \doi{10.1111/cgf.12184}.
\newblock URL \url{https://doi.org/10.1111/cgf.12184}.

\bibitem[Huang et~al.(2009)Huang, Wicke, Adams, and Guibas]{10.1111:j.1467-8659.2009.01380.x}
Qi-Xing Huang, Martin Wicke, Bart Adams, and Leonidas Guibas.
\newblock {Shape Decomposition using Modal Analysis}.
\newblock \emph{Computer Graphics Forum}, 28\penalty0 (2):\penalty0 407--416, 2009.
\newblock ISSN 1467-8659.
\newblock \doi{10.1111/j.1467-8659.2009.01380.x}.

\bibitem[Huang et~al.(2012)Huang, Zhang, Gao, Hu, Butscher, and Guibas]{10.1145/2366145.2366186}
Qi-Xing Huang, Guo-Xin Zhang, Lin Gao, Shi-Min Hu, Adrian Butscher, and Leonidas Guibas.
\newblock An optimization approach for extracting and encoding consistent maps in a shape collection.
\newblock \emph{ACM Trans. Graph.}, 31\penalty0 (6), nov 2012.
\newblock ISSN 0730-0301.
\newblock \doi{10.1145/2366145.2366186}.
\newblock URL \url{https://doi.org/10.1145/2366145.2366186}.

\bibitem[Huang et~al.(2008)Huang, Adams, Wicke, and Guibas]{qixing_2008_nonrigid}
Qixing Huang, Bart Adams, Martin Wicke, and Leonidas~J. Guibas.
\newblock Non-rigid registration under isometric deformations.
\newblock \emph{Comput. Graph. Forum}, 27\penalty0 (5):\penalty0 1449--1457, 2008.

\bibitem[Huang et~al.(2014)Huang, Wang, and Guibas]{DBLP:journals/tog/HuangWG14}
Qixing Huang, Fan Wang, and Leonidas~J. Guibas.
\newblock Functional map networks for analyzing and exploring large shape collections.
\newblock \emph{{ACM} Trans. Graph.}, 33\penalty0 (4):\penalty0 36:1--36:11, 2014.

\bibitem[Huang et~al.(2019{\natexlab{a}})Huang, Liang, Wang, Zuo, and Bajaj]{Huang:2019:Tensor:Maps}
Qixing Huang, Zhenxiao Liang, Haoyun Wang, Simiao Zuo, and Chandrajit Bajaj.
\newblock Tensor maps for synchronizing heterogeneous shape collections.
\newblock \emph{ACM Trans. Graph.}, 38\penalty0 (4), jul 2019{\natexlab{a}}.
\newblock ISSN 0730-0301.
\newblock \doi{10.1145/3306346.3322944}.
\newblock URL \url{https://doi.org/10.1145/3306346.3322944}.

\bibitem[Huang et~al.(2021)Huang, Huang, Sun, Zhang, Jiang, and Bajaj]{huang2021arapreg}
Qixing Huang, Xiangru Huang, Bo~Sun, Zaiwei Zhang, Junfeng Jiang, and Chandrajit Bajaj.
\newblock Arapreg: An as-rigid-as possible regularization loss for learning deformable shape generators.
\newblock In \emph{Proceedings of the IEEE/CVF international conference on computer vision}, pp.\  5815--5825, 2021.

\bibitem[Huang et~al.(2020{\natexlab{a}})Huang, Ren, Wonka, and Ovsjanikov]{Huang_2020_czo}
Ruqi Huang, Jing Ren, Peter Wonka, and Maks Ovsjanikov.
\newblock Consistent zoomout: Efficient spectral map synchronization.
\newblock \emph{Comput. Graph. Forum}, 39\penalty0 (5):\penalty0 265--278, 2020{\natexlab{a}}.

\bibitem[Huang et~al.(2017)Huang, Liang, Bajaj, and Huang]{NIPS2017_6744}
Xiangru Huang, Zhenxiao Liang, Chandrajit Bajaj, and Qixing Huang.
\newblock Translation synchronization via truncated least squares.
\newblock In I.~Guyon, U.~V. Luxburg, S.~Bengio, H.~Wallach, R.~Fergus, S.~Vishwanathan, and R.~Garnett (eds.), \emph{Advances in Neural Information Processing Systems 30}, pp.\  1459--1468. Curran Associates Inc., Red Hook, NY, USA, 2017.
\newblock URL \url{http://papers.nips.cc/paper/6744-translation-synchronization-via-truncated-least-squares.pdf}.

\bibitem[Huang et~al.(2019{\natexlab{b}})Huang, Liang, Zhou, Xie, Guibas, and Huang]{Huang_2019_CVPR}
Xiangru Huang, Zhenxiao Liang, Xiaowei Zhou, Yao Xie, Leonidas~J. Guibas, and Qixing Huang.
\newblock Learning transformation synchronization.
\newblock In \emph{The IEEE Conference on Computer Vision and Pattern Recognition (CVPR)}, pp.\  8082--8091, Virtual, 2019{\natexlab{b}}. Computer Vision Foundation / {IEEE}.

\bibitem[Huang et~al.(2020{\natexlab{b}})Huang, Liang, and Huang]{10.1145/3386569.3392402}
Xiangru Huang, Zhenxiao Liang, and Qixing Huang.
\newblock Uncertainty quantification for multi-scan registration.
\newblock \emph{ACM Trans. Graph.}, 39\penalty0 (4), July 2020{\natexlab{b}}.
\newblock ISSN 0730-0301.
\newblock \doi{10.1145/3386569.3392402}.
\newblock URL \url{https://doi.org/10.1145/3386569.3392402}.

\bibitem[Kim et~al.(2011)Kim, Lipman, and Funkhouser]{10.1145/2010324.1964974}
Vladimir~G. Kim, Yaron Lipman, and Thomas Funkhouser.
\newblock Blended intrinsic maps.
\newblock \emph{ACM Trans. Graph.}, 30\penalty0 (4), jul 2011.
\newblock ISSN 0730-0301.
\newblock \doi{10.1145/2010324.1964974}.
\newblock URL \url{https://doi.org/10.1145/2010324.1964974}.

\bibitem[Kim et~al.(2012)Kim, Li, Mitra, DiVerdi, and Funkhouser]{KIM:2012:EC}
Vladimir~G. Kim, Wilmot Li, Niloy~J. Mitra, Stephen DiVerdi, and Thomas Funkhouser.
\newblock Exploring collections of 3d models using fuzzy correspondences.
\newblock \emph{ACM Trans. Graph.}, 31\penalty0 (4), jul 2012.
\newblock ISSN 0730-0301.
\newblock \doi{10.1145/2185520.2185550}.
\newblock URL \url{https://doi.org/10.1145/2185520.2185550}.

\bibitem[Kraevoy \& Sheffer(2004)Kraevoy and Sheffer]{10.1145/1015706.1015811}
Vladislav Kraevoy and Alla Sheffer.
\newblock Cross-parameterization and compatible remeshing of 3d models.
\newblock \emph{ACM Trans. Graph.}, 23\penalty0 (3):\penalty0 861–869, aug 2004.
\newblock ISSN 0730-0301.
\newblock \doi{10.1145/1015706.1015811}.
\newblock URL \url{https://doi.org/10.1145/1015706.1015811}.

\bibitem[Li et~al.(2019{\natexlab{a}})Li, Zaheer, Zhang, P{\'{o}}czos, and Salakhutdinov]{DBLP:conf/iclr/LiZZPS19}
Chun{-}Liang Li, Manzil Zaheer, Yang Zhang, Barnab{\'{a}}s P{\'{o}}czos, and Ruslan Salakhutdinov.
\newblock Point cloud {GAN}, 2019{\natexlab{a}}.

\bibitem[Li et~al.(2022)Li, Donati, and Ovsjanikov]{DBLP:journals/corr/abs-2210-06373}
Lei Li, Nicolas Donati, and Maks Ovsjanikov.
\newblock Learning multi-resolution functional maps with spectral attention for robust shape matching.
\newblock In \emph{Advances in Neural Information Processing Systems 35: Annual Conference on Neural Information Processing Systems 2022, NeurIPS 2022, 8-14 December 2022, Vancouver, BC, Canada}, pp.\  1--10, 2022.

\bibitem[Li et~al.(2019{\natexlab{b}})Li, Li, Fu, Cohen{-}Or, and Heng]{DBLP:conf/iccv/LiLFCH19}
Ruihui Li, Xianzhi Li, Chi{-}Wing Fu, Daniel Cohen{-}Or, and Pheng{-}Ann Heng.
\newblock {PU-GAN:} {A} point cloud upsampling adversarial network.
\newblock In \emph{{ICCV}}, pp.\  7202--7211, Seoul, Republic of Korea, 2019{\natexlab{b}}. {IEEE}.

\bibitem[Litany et~al.(2017)Litany, Remez, Rodol{\`{a}}, Bronstein, and Bronstein]{DBLP:conf/iccv/LitanyRRBB17}
Or~Litany, Tal Remez, Emanuele Rodol{\`{a}}, Alexander~M. Bronstein, and Michael~M. Bronstein.
\newblock Deep functional maps: Structured prediction for dense shape correspondence.
\newblock In \emph{{ICCV}}, pp.\  5660--5668, Washington, DC, USA, 2017. {IEEE} Computer Society.

\bibitem[Litany et~al.(2018)Litany, Bronstein, Bronstein, and Makadia]{LitanyBBM18}
Or~Litany, Alexander~M. Bronstein, Michael~M. Bronstein, and Ameesh Makadia.
\newblock Deformable shape completion with graph convolutional autoencoders.
\newblock In \emph{2018 {IEEE} Conference on Computer Vision and Pattern Recognition, {CVPR} 2018, Salt Lake City, UT, USA, June 18-22, 2018}, pp.\  1886--1895, Salt Lake City, UT, USA, 2018. {IEEE} Computer Society.
\newblock \doi{10.1109/CVPR.2018.00202}.
\newblock URL \url{http://openaccess.thecvf.com/content\_cvpr\_2018/html/Litany\_Deformable\_Shape\_Completion\_CVPR\_2018\_paper.html}.

\bibitem[Lorensen \& Cline(1987)Lorensen and Cline]{DBLP:conf/siggraph/LorensenC87}
William~E. Lorensen and Harvey~E. Cline.
\newblock Marching cubes: {A} high resolution 3d surface construction algorithm.
\newblock In Maureen~C. Stone (ed.), \emph{Proceedings of the 14th Annual Conference on Computer Graphics and Interactive Techniques, {SIGGRAPH} 1987, Anaheim, California, USA, July 27-31, 1987}, pp.\  163--169, New York, NY, 1987. {ACM}.
\newblock \doi{10.1145/37401.37422}.
\newblock URL \url{https://doi.org/10.1145/37401.37422}.

\bibitem[Luo \& Hu(2021)Luo and Hu]{luo2021diffusion}
Shitong Luo and Wei Hu.
\newblock Diffusion probabilistic models for 3d point cloud generation.
\newblock In \emph{Proceedings of the IEEE/CVF Conference on Computer Vision and Pattern Recognition}, pp.\  2837--2845, 2021.

\bibitem[Maron et~al.(2016)Maron, Dym, Kezurer, Kovalsky, and Lipman]{DBLP:journals/tog/MaronDKKL16}
Haggai Maron, Nadav Dym, Itay Kezurer, Shahar~Z. Kovalsky, and Yaron Lipman.
\newblock Point registration via efficient convex relaxation.
\newblock \emph{{ACM} Trans. Graph.}, 35\penalty0 (4):\penalty0 73:1--73:12, 2016.

\bibitem[Melzi et~al.(2019)Melzi, Ren, Rodol{\`{a}}, Sharma, Wonka, and Ovsjanikov]{DBLP:journals/tog/MelziRRSWO19}
Simone Melzi, Jing Ren, Emanuele Rodol{\`{a}}, Abhishek Sharma, Peter Wonka, and Maks Ovsjanikov.
\newblock Zoomout: spectral upsampling for efficient shape correspondence.
\newblock \emph{{ACM} Trans. Graph.}, 38\penalty0 (6):\penalty0 155:1--155:14, 2019.
\newblock \doi{10.1145/3355089.3356524}.
\newblock URL \url{https://doi.org/10.1145/3355089.3356524}.

\bibitem[Mescheder et~al.(2019)Mescheder, Oechsle, Niemeyer, Nowozin, and Geiger]{DBLP:conf/cvpr/MeschederONNG19}
Lars~M. Mescheder, Michael Oechsle, Michael Niemeyer, Sebastian Nowozin, and Andreas Geiger.
\newblock Occupancy networks: Learning 3d reconstruction in function space.
\newblock In \emph{{CVPR}}, pp.\  4460--4470, Long Beach, California, 2019. Computer Vision Foundation / {IEEE}.

\bibitem[Muralikrishnan et~al.(2022)Muralikrishnan, Chaudhuri, Aigerman, Kim, Fisher, and Mitra]{Muralikrishnan_2022_glass}
Sanjeev Muralikrishnan, Siddhartha Chaudhuri, Noam Aigerman, Vladimir~G. Kim, Matthew Fisher, and Niloy~J. Mitra.
\newblock Glass: Geometric latent augmentation for shape spaces.
\newblock In \emph{Proceedings of the IEEE/CVF Conference on Computer Vision and Pattern Recognition (CVPR)}, pp.\  18552--18561, June 2022.

\bibitem[Nguyen et~al.(2011)Nguyen, Ben{-}Chen, Welnicka, Ye, and Guibas]{DBLP:journals/cgf/NguyenBWYG11}
Andy Nguyen, Mirela Ben{-}Chen, Katarzyna Welnicka, Yinyu Ye, and Leonidas~J. Guibas.
\newblock An optimization approach to improving collections of shape maps.
\newblock \emph{Comput. Graph. Forum}, 30\penalty0 (5):\penalty0 1481--1491, 2011.

\bibitem[Ovsjanikov et~al.(2012)Ovsjanikov, Ben-Chen, Solomon, Butscher, and Guibas]{Ovsjanikov:2012:FMAP}
Maks Ovsjanikov, Mirela Ben-Chen, Justin Solomon, Adrian Butscher, and Leonidas Guibas.
\newblock Functional maps: A flexible representation of maps between shapes.
\newblock \emph{ACM Trans. Graph.}, 31\penalty0 (4), jul 2012.
\newblock ISSN 0730-0301.
\newblock \doi{10.1145/2185520.2185526}.
\newblock URL \url{https://doi.org/10.1145/2185520.2185526}.

\bibitem[Park et~al.(2019)Park, Florence, Straub, Newcombe, and Lovegrove]{DBLP:conf/cvpr/ParkFSNL19}
Jeong~Joon Park, Peter Florence, Julian Straub, Richard~A. Newcombe, and Steven Lovegrove.
\newblock Deepsdf: Learning continuous signed distance functions for shape representation.
\newblock In \emph{{IEEE} Conference on Computer Vision and Pattern Recognition, {CVPR} 2019, Long Beach, CA, USA, June 16-20, 2019}, pp.\  165--174, Long Beach, California, 2019. Computer Vision Foundation / {IEEE}.
\newblock \doi{10.1109/CVPR.2019.00025}.
\newblock URL \url{http://openaccess.thecvf.com/content\_CVPR\_2019/html/Park\_DeepSDF\_Learning\_Continuous\_Signed\_Distance\_Functions\_for\_Shape\_Representation\_CVPR\_2019\_paper.html}.

\bibitem[Paszke et~al.(2019)Paszke, Gross, Massa, Lerer, Bradbury, Chanan, Killeen, Lin, Gimelshein, Antiga, Desmaison, K{\"{o}}pf, Yang, DeVito, Raison, Tejani, Chilamkurthy, Steiner, Fang, Bai, and Chintala]{DBLP:conf/nips/PaszkeGMLBCKLGA19}
Adam Paszke, Sam Gross, Francisco Massa, Adam Lerer, James Bradbury, Gregory Chanan, Trevor Killeen, Zeming Lin, Natalia Gimelshein, Luca Antiga, Alban Desmaison, Andreas K{\"{o}}pf, Edward~Z. Yang, Zachary DeVito, Martin Raison, Alykhan Tejani, Sasank Chilamkurthy, Benoit Steiner, Lu~Fang, Junjie Bai, and Soumith Chintala.
\newblock Pytorch: An imperative style, high-performance deep learning library.
\newblock In \emph{Annual Conference on Neural Information Processing Systems 2019, NeurIPS 2019, December 8-14, 2019, Vancouver, BC, Canada}, pp.\  8024--8035, 2019.

\bibitem[Peng et~al.(2021)Peng, Zhang, Xu, Wang, Shuai, Bao, and Zhou]{DBLP:conf/cvpr/PengZXWSBZ21}
Sida Peng, Yuanqing Zhang, Yinghao Xu, Qianqian Wang, Qing Shuai, Hujun Bao, and Xiaowei Zhou.
\newblock Neural body: Implicit neural representations with structured latent codes for novel view synthesis of dynamic humans.
\newblock In \emph{{CVPR}}, pp.\  9054--9063, Washington, DC, 2021. Computer Vision Foundation / {IEEE}.

\bibitem[Qi et~al.(2017)Qi, Su, Mo, and Guibas]{Qi_2017_pointnet}
Charles~R. Qi, Hao Su, Kaichun Mo, and Leonidas~J. Guibas.
\newblock Pointnet: Deep learning on point sets for 3d classification and segmentation.
\newblock In \emph{Proceedings of the IEEE Conference on Computer Vision and Pattern Recognition (CVPR)}, July 2017.

\bibitem[Rakotosaona \& Ovsjanikov(2020)Rakotosaona and Ovsjanikov]{RakotosaonaO20}
Marie{-}Julie Rakotosaona and Maks Ovsjanikov.
\newblock Intrinsic point cloud interpolation via dual latent space navigation.
\newblock In Andrea Vedaldi, Horst Bischof, Thomas Brox, and Jan{-}Michael Frahm (eds.), \emph{Computer Vision - {ECCV} 2020 - 16th European Conference, Glasgow, UK, August 23-28, 2020, Proceedings, Part {II}}, volume 12347 of \emph{Lecture Notes in Computer Science}, pp.\  655--672, New York, NY, 2020. Springer.
\newblock \doi{10.1007/978-3-030-58536-5\_39}.
\newblock URL \url{https://doi.org/10.1007/978-3-030-58536-5\_39}.

\bibitem[Ranjan et~al.(2018)Ranjan, Bolkart, Sanyal, and Black]{ranjan2018generating}
Anurag Ranjan, Timo Bolkart, Soubhik Sanyal, and Michael~J. Black.
\newblock Generating 3d faces using convolutional mesh autoencoders.
\newblock In Vittorio Ferrari, Martial Hebert, Cristian Sminchisescu, and Yair Weiss (eds.), \emph{Computer Vision - {ECCV} 2018 - 15th European Conference, Munich, Germany, September 8-14, 2018, Proceedings, Part {III}}, volume 11207 of \emph{Lecture Notes in Computer Science}, pp.\  725--741, New York, NY, 2018. Springer.
\newblock \doi{10.1007/978-3-030-01219-9\_43}.
\newblock URL \url{https://doi.org/10.1007/978-3-030-01219-9\_43}.

\bibitem[Ren et~al.(2018)Ren, Poulenard, Wonka, and Ovsjanikov]{Ren:2018:COFM}
Jing Ren, Adrien Poulenard, Peter Wonka, and Maks Ovsjanikov.
\newblock Continuous and orientation-preserving correspondences via functional maps.
\newblock \emph{ACM Trans. Graph.}, 37\penalty0 (6), dec 2018.
\newblock ISSN 0730-0301.
\newblock \doi{10.1145/3272127.3275040}.
\newblock URL \url{https://doi.org/10.1145/3272127.3275040}.

\bibitem[Sahillioglu(2020)]{DBLP:journals/vc/Sahillioglu20}
Yusuf Sahillioglu.
\newblock Recent advances in shape correspondence.
\newblock \emph{Vis. Comput.}, 36\penalty0 (8):\penalty0 1705--1721, 2020.

\bibitem[Saito et~al.(2019)Saito, Huang, Natsume, Morishima, Li, and Kanazawa]{DBLP:conf/iccv/SaitoHNMLK19}
Shunsuke Saito, Zeng Huang, Ryota Natsume, Shigeo Morishima, Hao Li, and Angjoo Kanazawa.
\newblock Pifu: Pixel-aligned implicit function for high-resolution clothed human digitization.
\newblock In \emph{{ICCV}}, pp.\  2304--2314, Seoul, Republic of Korea, 2019. {IEEE}.

\bibitem[Saito et~al.(2020)Saito, Simon, Saragih, and Joo]{DBLP:conf/cvpr/SaitoSSJ20}
Shunsuke Saito, Tomas Simon, Jason~M. Saragih, and Hanbyul Joo.
\newblock Pifuhd: Multi-level pixel-aligned implicit function for high-resolution 3d human digitization.
\newblock In \emph{{CVPR}}, pp.\  81--90, Washington, DC, 2020. Computer Vision Foundation / {IEEE}.

\bibitem[Schreiner et~al.(2004)Schreiner, Asirvatham, Praun, and Hoppe]{10.1145/1186562.1015812}
John Schreiner, Arul Asirvatham, Emil Praun, and Hugues Hoppe.
\newblock Inter-surface mapping.
\newblock In \emph{ACM SIGGRAPH 2004 Papers}, SIGGRAPH '04, pp.\  870–877, New York, NY, USA, 2004. Association for Computing Machinery.
\newblock ISBN 9781450378239.
\newblock \doi{10.1145/1186562.1015812}.
\newblock URL \url{https://doi.org/10.1145/1186562.1015812}.

\bibitem[Sharma \& Ovsjanikov(2020)Sharma and Ovsjanikov]{SharmaO20-0}
Abhishek Sharma and Maks Ovsjanikov.
\newblock Weakly supervised deep functional maps for shape matching.
\newblock In Hugo Larochelle, Marc'Aurelio Ranzato, Raia Hadsell, Maria{-}Florina Balcan, and Hsuan{-}Tien Lin (eds.), \emph{Advances in Neural Information Processing Systems 33: Annual Conference on Neural Information Processing Systems 2020, NeurIPS 2020, December 6-12, 2020, virtual}, pp.\  19264--19275, Red Hook, NY, USA, 2020. Curran Associates Inc.
\newblock URL \url{https://proceedings.neurips.cc/paper/2020/hash/dfb84a11f431c62436cfb760e30a34fe-Abstract.html}.

\bibitem[Slavcheva et~al.(2017)Slavcheva, Baust, Cremers, and Ilic]{SlavchevaBCI17}
Miroslava Slavcheva, Maximilian Baust, Daniel Cremers, and Slobodan Ilic.
\newblock Killingfusion: Non-rigid 3d reconstruction without correspondences.
\newblock In \emph{2017 {IEEE} Conference on Computer Vision and Pattern Recognition, {CVPR} 2017, Honolulu, HI, USA, July 21-26, 2017}, pp.\  5474--5483, Washington, DC, 2017. {IEEE} Computer Society.
\newblock \doi{10.1109/CVPR.2017.581}.
\newblock URL \url{https://doi.org/10.1109/CVPR.2017.581}.

\bibitem[Solomon et~al.(2011)Solomon, Ben{-}Chen, Butscher, and Guibas]{DBLP:journals/cgf/SolomonBBG11a}
Justin Solomon, Mirela Ben{-}Chen, Adrian Butscher, and Leonidas~J. Guibas.
\newblock As-killing-as-possible vector fields for planar deformation.
\newblock \emph{Comput. Graph. Forum}, 30\penalty0 (5):\penalty0 1543--1552, 2011.

\bibitem[Sorkine \& Alexa(2007)Sorkine and Alexa]{olga_2007_arap}
Olga Sorkine and Marc Alexa.
\newblock As-rigid-as-possible surface modeling.
\newblock In \emph{Proceedings of the Fifth Eurographics Symposium on Geometry Processing, Barcelona, Spain, July 4-6, 2007}, volume 257, pp.\  109--116, 2007.

\bibitem[Stam \& Schmidt(2011)Stam and Schmidt]{DBLP:journals/tog/StamS11}
Jos Stam and Ryan~M. Schmidt.
\newblock On the velocity of an implicit surface.
\newblock \emph{{ACM} Trans. Graph.}, 30\penalty0 (3):\penalty0 21:1--21:7, 2011.

\bibitem[Sumner \& Popovi{\'c}(2004)Sumner and Popovi{\'c}]{sumner2004deformation}
Robert~W Sumner and Jovan Popovi{\'c}.
\newblock Deformation transfer for triangle meshes.
\newblock \emph{ACM Transactions on graphics (TOG)}, 23\penalty0 (3):\penalty0 399--405, 2004.

\bibitem[Sundararaman et~al.(2022)Sundararaman, Pai, and Ovsjanikov]{DBLP:conf/eccv/SundararamanPO22}
Ramana Sundararaman, Gautam Pai, and Maks Ovsjanikov.
\newblock Implicit field supervision for robust non-rigid shape matching.
\newblock In Shai Avidan, Gabriel~J. Brostow, Moustapha Ciss{\'{e}}, Giovanni~Maria Farinella, and Tal Hassner (eds.), \emph{Computer Vision - {ECCV} 2022 - 17th European Conference, Tel Aviv, Israel, October 23-27, 2022, Proceedings, Part {III}}, volume 13663 of \emph{Lecture Notes in Computer Science}, pp.\  344--362. Springer, 2022.
\newblock \doi{10.1007/978-3-031-20062-5\_20}.
\newblock URL \url{https://doi.org/10.1007/978-3-031-20062-5\_20}.

\bibitem[Tan et~al.(2018)Tan, Gao, Lai, and Xia]{Tan0LX18}
Qingyang Tan, Lin Gao, Yu{-}Kun Lai, and Shihong Xia.
\newblock Variational autoencoders for deforming 3d mesh models.
\newblock In \emph{2018 {IEEE} Conference on Computer Vision and Pattern Recognition, {CVPR} 2018, Salt Lake City, UT, USA, June 18-22, 2018}, pp.\  5841--5850, Salt Lake City, UT, USA, 2018. {IEEE} Computer Society.
\newblock \doi{10.1109/CVPR.2018.00612}.
\newblock URL \url{http://openaccess.thecvf.com/content\_cvpr\_2018/html/Tan\_Variational\_Autoencoders\_for\_CVPR\_2018\_paper.html}.

\bibitem[Tao et~al.(2016)Tao, Solomon, and Butscher]{DBLP:journals/cgf/TaoSB16}
Michael Tao, Justin Solomon, and Adrian Butscher.
\newblock Near-isometric level set tracking.
\newblock \emph{Comput. Graph. Forum}, 35\penalty0 (5):\penalty0 65--77, 2016.

\bibitem[Tretschk et~al.(2020)Tretschk, Tewari, Zollh{\"{o}}fer, Golyanik, and Theobalt]{TretschkTZGT20}
Edgar Tretschk, Ayush Tewari, Michael Zollh{\"{o}}fer, Vladislav Golyanik, and Christian Theobalt.
\newblock {DEMEA:} deep mesh autoencoders for non-rigidly deforming objects.
\newblock In Andrea Vedaldi, Horst Bischof, Thomas Brox, and Jan{-}Michael Frahm (eds.), \emph{Computer Vision - {ECCV} 2020 - 16th European Conference, Glasgow, UK, August 23-28, 2020, Proceedings, Part {IV}}, volume 12349 of \emph{Lecture Notes in Computer Science}, pp.\  601--617, Glasgow,UK, 2020. Springer.
\newblock \doi{10.1007/978-3-030-58548-8\_35}.
\newblock URL \url{https://doi.org/10.1007/978-3-030-58548-8\_35}.

\bibitem[van Kaick et~al.(2010)van Kaick, Zhang, Hamarneh, and Cohen{-}Or]{DBLP:conf/eurographics/Kaick0HC10}
Oliver van Kaick, Hao Zhang, Ghassan Hamarneh, and Daniel Cohen{-}Or.
\newblock A survey on shape correspondence.
\newblock In Helwig Hauser and Erik Reinhard (eds.), \emph{31st Annual Conference of the European Association for Computer Graphics, Eurographics 2010 - State of the Art Reports, Norrk{\"{o}}ping, Sweden, May 3-7, 2010}, pp.\  61--82, 9105 Salley Street Montreal, Quebec H8R 2C8 CANADA, 2010. Eurographics Association.
\newblock \doi{10.2312/egst.20101062}.
\newblock URL \url{https://doi.org/10.2312/egst.20101062}.

\bibitem[Verma et~al.(2018)Verma, Boyer, and Verbeek]{DBLP:conf/cvpr/VermaBV18}
Nitika Verma, Edmond Boyer, and Jakob Verbeek.
\newblock Feastnet: Feature-steered graph convolutions for 3d shape analysis.
\newblock In \emph{2018 {IEEE} Conference on Computer Vision and Pattern Recognition, {CVPR} 2018, Salt Lake City, UT, USA, June 18-22, 2018}, pp.\  2598--2606, Salt Lake City, UT, USA, 2018. {IEEE} Computer Society.
\newblock \doi{10.1109/CVPR.2018.00275}.
\newblock URL \url{http://openaccess.thecvf.com/content\_cvpr\_2018/html/Verma\_FeaStNet\_Feature-Steered\_Graph\_CVPR\_2018\_paper.html}.

\bibitem[Wang \& Singer(2013)Wang and Singer]{Wang:2013:IMA}
Lanhui Wang and Amit Singer.
\newblock Exact and stable recovery of rotations for robust synchronization.
\newblock \emph{Information and Inference: A Journal of the IMA}, 2:\penalty0 145–193, December 2013.

\bibitem[Yang et~al.(2018)Yang, Feng, Shen, and Tian]{DBLP:conf/cvpr/YangFST18}
Yaoqing Yang, Chen Feng, Yiru Shen, and Dong Tian.
\newblock Foldingnet: Point cloud auto-encoder via deep grid deformation.
\newblock In \emph{{CVPR}}, pp.\  206--215, Washington, DC, 2018. Computer Vision Foundation / {IEEE} Computer Society.

\bibitem[Yoshiyasu et~al.(2014)Yoshiyasu, Ma, Yoshida, and Kanehiro]{10.1111/cgf.12451}
Yusuke Yoshiyasu, Wan-Chun Ma, Eiichi Yoshida, and Fumio Kanehiro.
\newblock As-conformal-as-possible surface registration.
\newblock In \emph{Proceedings of the Symposium on Geometry Processing}, SGP '14, pp.\  257–267, Goslar, DEU, 2014. Eurographics Association.
\newblock \doi{10.1111/cgf.12451}.
\newblock URL \url{https://doi.org/10.1111/cgf.12451}.

\bibitem[Zhou et~al.(2020)Zhou, Wu, Li, Cao, Ye, Saragih, Li, and Sheikh]{zhou2020fully}
Yi~Zhou, Chenglei Wu, Zimo Li, Chen Cao, Yuting Ye, Jason~M. Saragih, Hao Li, and Yaser Sheikh.
\newblock Fully convolutional mesh autoencoder using efficient spatially varying kernels.
\newblock In Hugo Larochelle, Marc'Aurelio Ranzato, Raia Hadsell, Maria{-}Florina Balcan, and Hsuan{-}Tien Lin (eds.), \emph{Advances in Neural Information Processing Systems 33: Annual Conference on Neural Information Processing Systems 2020, NeurIPS 2020, December 6-12, 2020, virtual}, pp.\  9251--9262, Red Hook, NY, USA, 2020. Curran Associates Inc.
\newblock URL \url{https://proceedings.neurips.cc/paper/2020/hash/68dd09b9ff11f0df5624a690fe0f6729-Abstract.html}.

\bibitem[Zuffi et~al.(2017)Zuffi, Kanazawa, Jacobs, and Black]{zuffi20173d}
Silvia Zuffi, Angjoo Kanazawa, David~W. Jacobs, and Michael~J. Black.
\newblock 3d menagerie: Modeling the 3d shape and pose of animals.
\newblock In \emph{{CVPR}}, pp.\  5524--5532, Washington, DC, USA, 2017. {IEEE} Computer Society.

\end{thebibliography}
\bibliographystyle{iclr2024_conference}

\appendix
\clearpage
\appendixpage

The supplementary materials provide more details of implementing the regularization loss used in the implicit shape generator in Section~\ref{Supp:Reg} and details of mesh generator initialization in Section~\ref{Supp:Mesh:Init}. Section~\ref{Supp:Datasets} gives more details on the dataset preprocessing. Section~\ref{Supp:Results:Shape_space} and Section~\ref{Supp:Results:Corres} show more results on shape space learning and shape matching, respectively.

\section{Details of Regularization Loss}
\label{Supp:Reg}

\subsection{Expression of $\overline{L}^{\arap}(\bs{g}^{\phi}(\bs{z}))$}
\label{Supp:Reg:Exp:ARAP}
\begin{align*}
\overline{L}^{\arap}(\bs{g}^{\phi}(\bs{z})) = & 2L\otimes I_3 - B^{\arap}\big(\bs{g}^{\phi}(\bs{z})\big)D^{\arap}\big(\bs{g}^{\phi}(\bs{z})\big){B^{\arap}\big(\bs{g}^{\phi}(\bs{z})\big)}^T,
\end{align*}
where $L$ is the graph Laplacian of the mesh, and $B^{\arap}\big(\bs{g}^{\phi}(\bs{z})\big)\in \R^{3n\times 3n}$ is a sparse block matrix defined as
$$
B^{\arap}_{ij}\big(\bs{g}^{\phi}(\bs{z})\big)  = \left\{
\begin{array}{cc}
\sum\limits_{k\in \set{N}_i}\bs{e}_{ik}^{\phi}(\bs{z}) \times  & i = j \\
\bs{e}_{ij}^{\phi}(\bs{z}) \times  & j\in \set{N}_i \\
 0 & \textup{else}
\end{array}
\right. 
$$
where $\bs{e}_{ij}^{\phi}(\bs{z}) =\bs{g}_i^{\phi}(\bs{z})-\bs{g}_j^{\phi}(\bs{z})$ and $D^{\arap}\big(\bs{g}^{\phi}(\bs{z})\big)\in \R^{3n\times 3n}$ is a diagonal block matrix defined as 
$$
D^{\arap}_{ii}\big(\bs{g}^{\phi}(\bs{z})\big)  =
\Big(\sum\limits_{j\in \set{N}_i}\big(\|\bs{e}_{ij}^{\phi}(\bs{z}) \|^2 I_3-\bs{e}_{ij}^{\phi}(\bs{z}) {\bs{e}_{ij}^{\phi}(\bs{z})} ^T\big)\Big)^{-1}
$$

\subsection{Expression of $\overline{L}^{\acap}(\bs{g}^{\phi}(\bs{z}))$}
\label{Supp:Reg:Exp:ACAP}

\begin{align*}
\overline{L}^{\acap}(\bs{g}^{\phi}(\bs{z})) = & 2L\otimes I_3 - B^{\acap}\big(\bs{g}^{\phi}(\bs{z})\big)D^{\acap}\big(\bs{g}^{\phi}(\bs{z})\big){B^{\acap}\big(\bs{g}^{\phi}(\bs{z})\big)}^T,
\end{align*}
where $B^{\acap}\big(\bs{g}^{\phi}(\bs{z})\big)\in \R^{3n\times 4n}$ is a sparse block matrix defined as
$$
B^{\acap}_{ij}\big(\bs{g}^{\phi}(\bs{z})\big)  = \left\{
\begin{array}{cc}
\sum\limits_{k\in \set{N}_i}\big(\begin{array}{cc}
-\bs{e}_{ik}^{\phi}(\bs{z}) & \bs{e}_{ik}^{\phi}(\bs{z}) \times
\end{array}\big)  & i = j \\
\big(\begin{array}{cc}
-\bs{e}_{ij}^{\phi}(\bs{z}) & \bs{e}_{ij}^{\phi}(\bs{z}) \times
\end{array}\big)
&j\in \set{N}_i \\
 0 & \textup{else}
\end{array}
\right. 
$$
and $D^{\acap}\big(\bs{g}^{\phi}(\bs{z})\big)\in \R^{4n\times 4n}$ is a diagonal block matrix defined as 
\begin{align*}
D^{\acap}_{ii}\big(\bs{g}^{\phi}(\bs{z})\big)  =
\Big(&\sum\limits_{j\in \set{N}_i}\big(\|\bs{e}_{ij}^{\phi}(\bs{z}) \|^2 I_4 - \diag(0,\bs{e}_{ij}^{\phi}(\bs{z}) {\bs{e}_{ij}^{\phi}(\bs{z})} ^T\big) \Big)^{-1}
\end{align*}

\subsection{Expression of $C^{\phi}(\bs{z})$ and $F^{\phi}(\bs{z})$}

Considering all vertices, the matrix representation of 
\begin{equation}
\frac{\partial f^{\phi}}{\partial \bs{x}}(\bs{g}_i^{\phi}(\bs{z}),\bs{z})^T\bs{d}_i^{\bs{v}}(\bs{z}) +\epsilon\frac{\partial f^{\phi}}{\partial \bs{z}}(\bs{g}_i^{\phi}(\bs{z}),\bs{z})^T\bs{v} = 0. \nonumber
\end{equation}
can be written as $C^{\phi}(\bs{z})\bs{d} = -\epsilon F^{\phi}(\bs{z})\bs{v}$, where
\begin{align*}
    \bs{d} = \left[ \begin{array}{c}
    \bs{d}_1^{\bs{v}}(\bs{z}) \\
    \bs{d}_2^{\bs{v}}(\bs{z}) \\
    \vdots \\
    \bs{d}_n^{\bs{v}}(\bs{z})
    \end{array} \right] \in \R^{3n},
\end{align*}

\begin{align*}
    C^{\phi}(\bs{z}) = \left[ \begin{array}{cccc}
    \frac{\partial f^{\phi}}{\partial \bs{x}}(\bs{g}_1^{\phi}(\bs{z}),\bs{z})^T & & & \\
    & \frac{\partial f^{\phi}}{\partial \bs{x}}(\bs{g}_2^{\phi}(\bs{z}),\bs{z})^T & & \\
    & & \hdots & \\
    & & & \frac{\partial f^{\phi}}{\partial \bs{x}}(\bs{g}_n^{\phi}(\bs{z}),\bs{z})^T
    \end{array} \right] \in \R^{n\times 3n},
\end{align*}

\begin{align*}
    F^{\phi}(\bs{z}) = \left[ \begin{array}{c}
    \frac{\partial f^{\phi}}{\partial \bs{z}}(\bs{g}_1^{\phi}(\bs{z}),\bs{z})^T \\
    \frac{\partial f^{\phi}}{\partial \bs{z}}(\bs{g}_2^{\phi}(\bs{z}),\bs{z})^T \\
    \vdots \\
    \frac{\partial f^{\phi}}{\partial \bs{z}}(\bs{g}_n^{\phi}(\bs{z}),\bs{z})^T
    \end{array} \right] \in \R^{n\times d},
\end{align*}

\subsection{Implementation Details}
\label{Supp:Reg:Implementation}

%

%

Both the geometric deformation regularization $r_{\geo}(f^{\phi}) $ and the cycle-consistency regularization $r_{\cycle}(f^{\phi})$ rely on the mesh with $n$ vertices that is discretized from $f^{\phi}(\bs x, \bs z) = 0$. We use Marching Cube for discretization. For the human category, we use a voxel grid with size $64\times 77\times64$. For the animal category, the size of the voxel grid is $82 \times 50 \times 71$. The output mesh from the Marching Cube algorithm typically contains more than 5000 vertices. To reduce the computation complexity, we simplify the output mesh into 2000 faces~\cite{GarlandH_1997_sim} before computing  $r_{\geo}(f^{\phi}) $ and $r_{\cycle}(f^{\phi})$. The number of vertices $n$ is around 1000, thus the size of $\overline{L}^{\phi}(\bs{z})$ is around $3000\times 3000$. Computing $(\overline{L}^{\phi}(\bs{z}))^{+}$ only takes about $40$ms in PyTorch~\cite{DBLP:conf/nips/PaszkeGMLBCKLGA19}
\section{Details of Mesh Generator Initialization}
\label{Supp:Mesh:Init}

\subsection{Template-based Registration}
In order to register the template mesh $\set{M}$ to the input shape $S_i$, we first generate $T$ intermediate shape $\bs{g}^{\phi}(\bs{z}_i^{j})$, where $\bs{z}_i^j = \bs{z}_{\temp} + j\frac{\bs{z}_i-\bs{z}_{\temp}}{T+1}, 1\leq j \leq T$. Instead of directly register $\set{M}$ to $S_i$, we first register $\set{M}$ to $\bs{g}^{\phi}(\bs{z}_i^{1})$ with ARAP deformation energy~\cite{olga_2007_arap, qixing_2008_nonrigid}. Since $\set{M}$ and $\bs{g}^{\phi}(\bs{z}_i^{1})$ are very close, we directly apply nearest neighbor search to compute the correspondence for the data term. The registration gives the resulting deformed template $\set{M}_1$. We then register $\set{M}_1$ to $\bs{g}^{\phi}(\bs{z}_i^{2})$ and get the deformed template $\set{M}_2$, register $\set{M}_2$ to $\bs{g}^{\phi}(\bs{z}_i^{3})$ and get the deformed template $\set{M}_3$, and so on and so forth. Finally we get the deformed template $\set{M}_m$, which is well-aligned with $S_i$.

The interpolation-guided registration typically works well but might fail when the template $\set{M}$ is too far from $S_i$, i.e. the two shapes have very different poses. The reason is that the intermediate shapes on the interpolation path might not have good quality. The make full use of the learned shape space, we add shapes from the input shape collection to the interpolation path in these cases. First, we compute the distance between each pair of shape $S_i$ and $S_j$ using the distance of their embedded latent codes $\|\bs{z}_i - \bs{z}_j\|$. Based on this distance metric, we build a $K$-NN graph among the input shape collection. We set $K=25$ for the human dataset and $K=40$ for the animal dataset. We then perform interpolation-guided registration on each edge $(i, j)$ of the graph and obtain the correspondences between $S_i$ and $S_j$. We use the distortion of the mapped edges~\cite{qixing_2008_nonrigid} as the weights in the $K$-NN graph. Finally we compute the shortest path from the template to each shape $S_i$ and get the correspondences by composing the correspondences along the shortest path.


\subsection{Mesh Generator Architecture}

The network architecture of $\bs{m}^{\theta}$ follows from that in~\cite{huang2021arapreg}, which outputs displacements of vertex positions of the template mesh $\set{M}$. We sample 4 resolutions of the mesh connections of the template mesh.  The network architecture stacks 4 blocks of convolution + up-sampling layers. The convolution layer employs Chebyshev convolutional filters with $6$ Chebyshev polynomials~\cite{ranjan2018generating}. Similar to~\cite{zhou2020fully}, there is a fully connected layer between the latent code and the input to the first convolution layer.

\section{Details of Datasets}
\label{Supp:Datasets}
\begin{figure}
\centering
\includegraphics[width=0.5\textwidth]{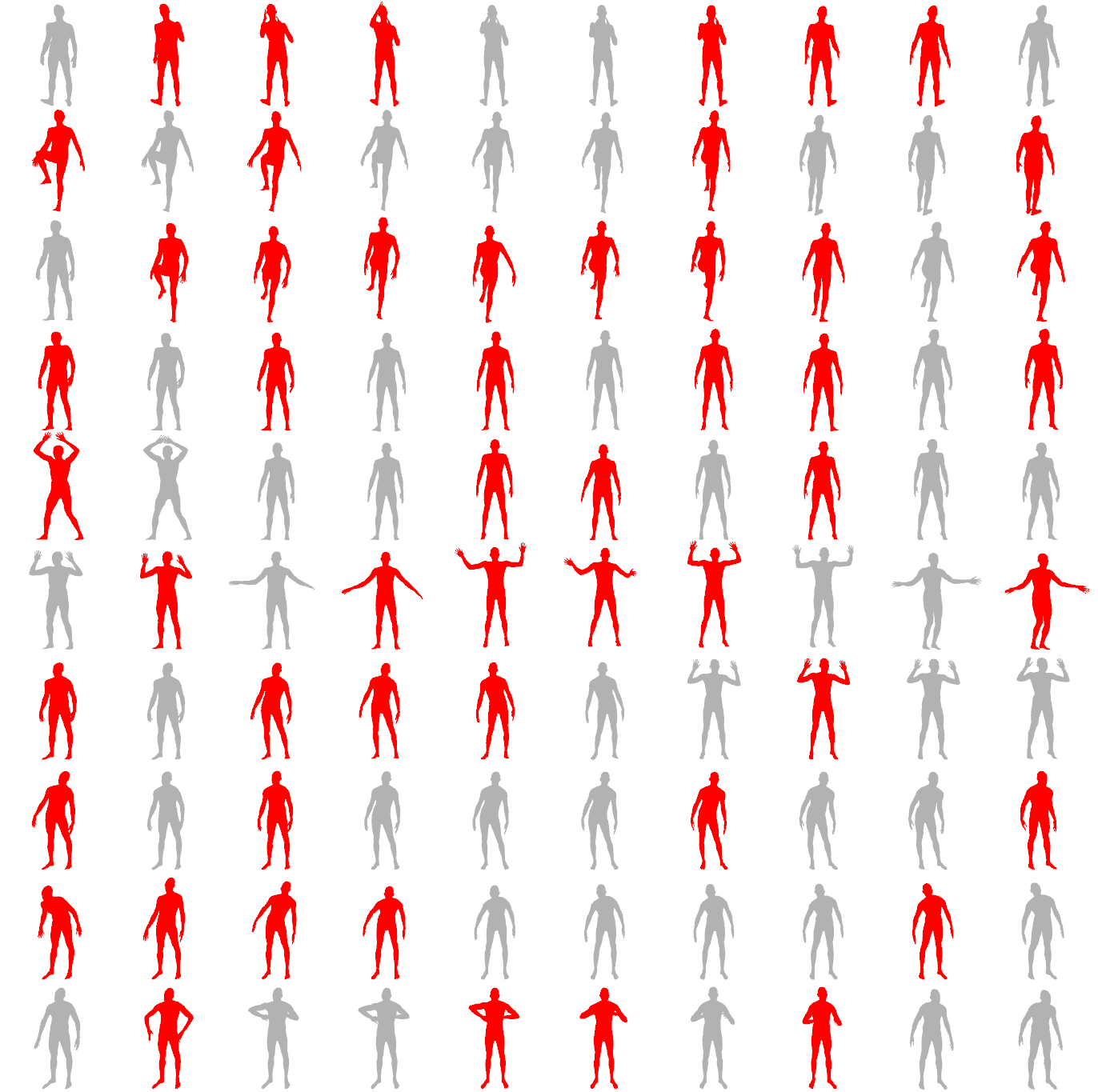}
\caption{ We use FPS to further select a more diverse and challenging subset of 1000 shapes from the training split~\cite{atzmon_2021_sald, Atzmon_2020_sal, gropp_2020_igr}. In this example, among $10\times10=100$ shapes from the training split~\cite{atzmon_2021_sald, Atzmon_2020_sal, gropp_2020_igr}, only the red shapes are selected because the gray shapes have similar poses.  }
\label{fig:supp:select} 
\end{figure}

There are approximately 41k shapes in the original DFAUST dataset. Since there is low variety between the adjacent shapes, recent works~\cite{atzmon_2021_sald, Atzmon_2020_sal, gropp_2020_igr} create a new training/testing split by uniformly sample $20\%$ shapes from the original dataset. However, we notice that there are still many similar shapes in the splits. For example, almost all motion sequences start from shapes with very similar rest pose. In order to make the dataset more challenging, we further select 1000 shapes from the training split. Specifically, we first learn a VAE~\cite{atzmon_2021_sald} to embed all the shapes from the training split into different latent vectors. We find that the latent vectors of similar shapes are typically closer. Then we apply farthest point sampling (FPS) to the latent vectors and select the first 1000 shapes. An example is shown in Figure~\ref{fig:supp:select}. We apply the same approach to the testing split and select another 1000 shapes for testing.

The original SMAL dataset from~\cite{huang2021arapreg} contains 300 training shapes and 100 testing shapes. We filter out the shapes that have unreasonable self-intersection, leading to 289 training shapes and 94 testing shapes.

For both DFAUST and SMAL dataset, we evaluate the correspondences from a template shape to the remaining shapes.

\section{More Results of Shape Space Learning}
\label{Supp:Results:Shape_space}
\begin{figure*}
\centering
\begin{overpic}[width=1.0\textwidth]{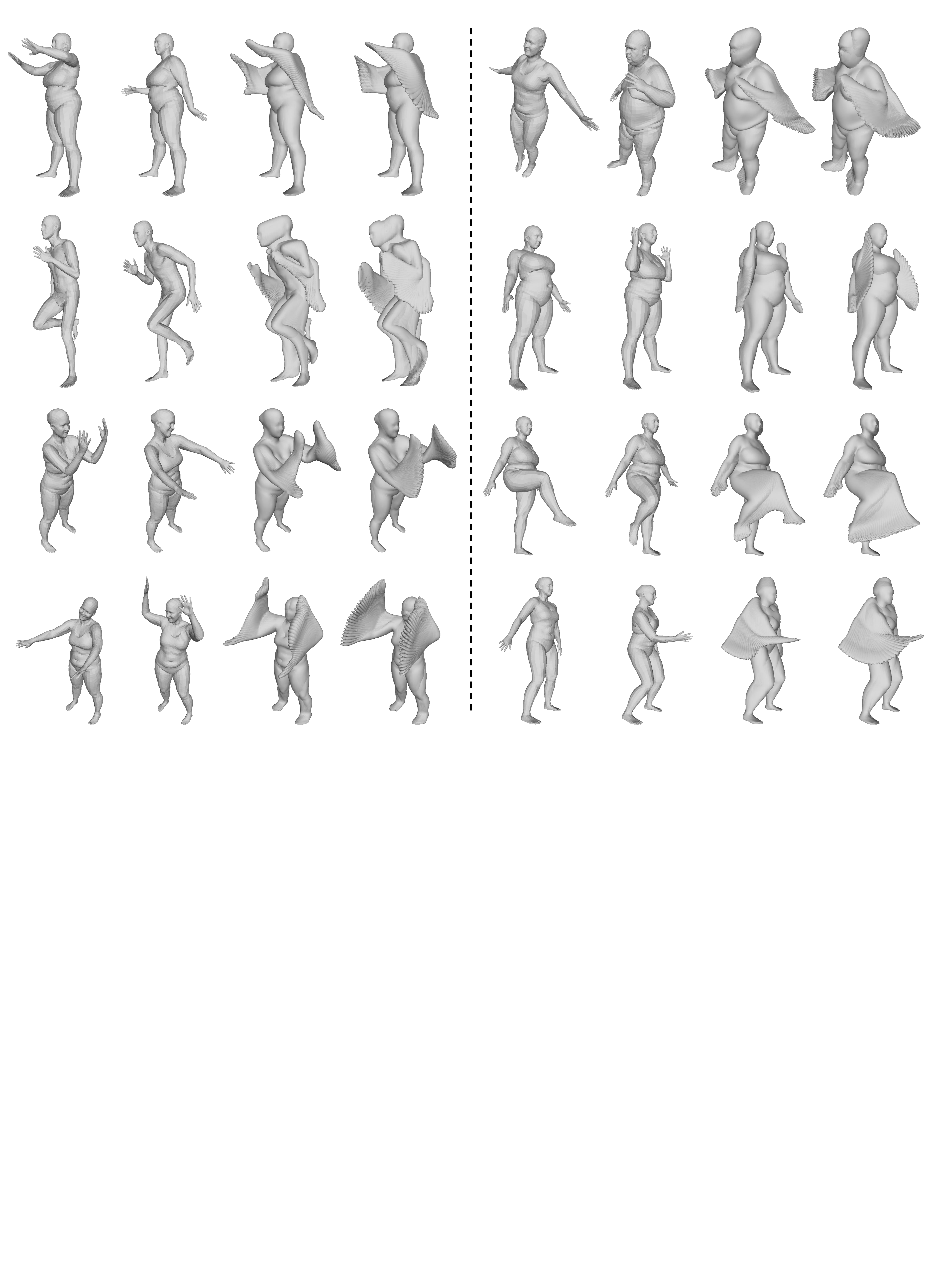}
\put(4 , -1){(a)}
\put(16, -1){(b)}
\put(27, -1){(c)}
\put(40, -1){(d)}
\put(57, -1){(a)}
\put(68, -1){(b)}
\put(80, -1){(c)}
\put(92, -1){(d)}
\end{overpic}
\caption{The comparison of shape interpolation between SALD~\cite{atzmon_2021_sald} and our method on the DFAUST dataset. (a) source shape. (b) target shape. (c) interpolation results of SALD~\cite{atzmon_2021_sald}. (d) our results. }
\label{fig:supp:shape_space:dfaust} 
\end{figure*}

\begin{figure*}
\centering
\begin{overpic}[width=1.0\textwidth]{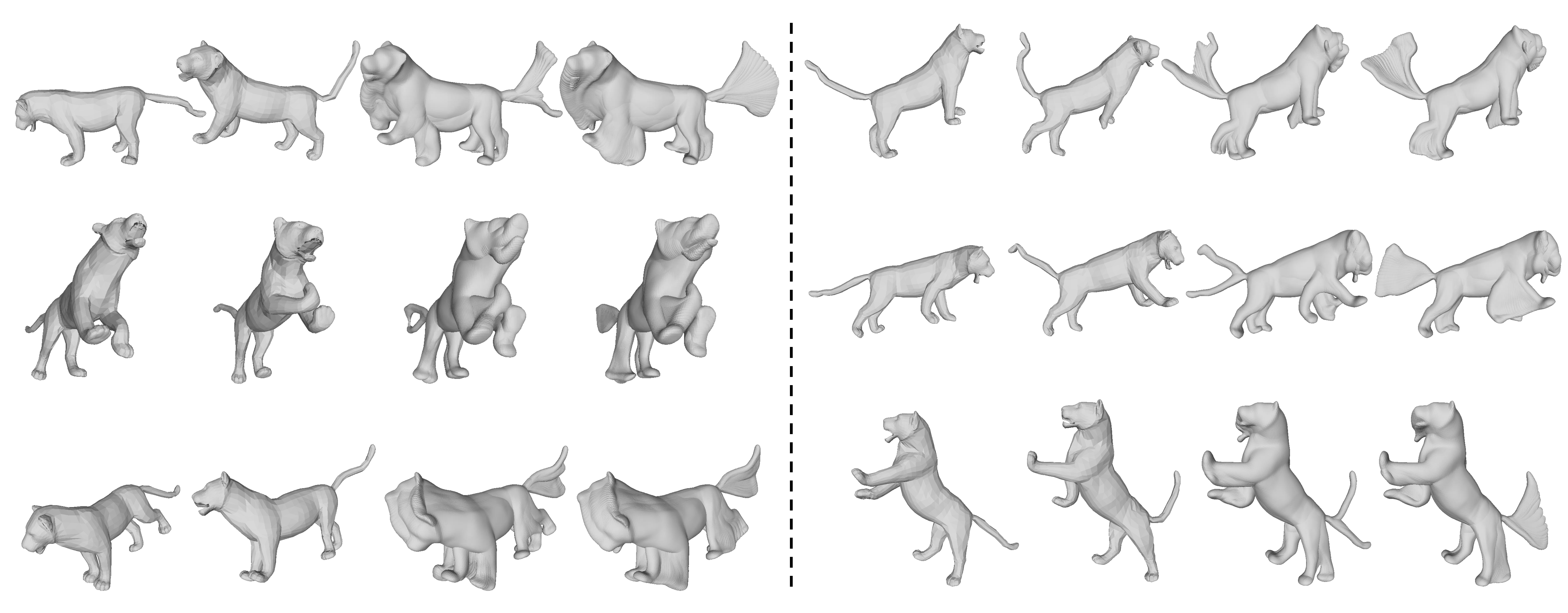}
\put(4 , -1){(a)}
\put(15, -1){(b)}
\put(29, -1){(c)}
\put(40, -1){(d)}
\put(57, -1){(a)}
\put(69, -1){(b)}
\put(80, -1){(c)}
\put(92, -1){(d)}
\end{overpic}
\vspace{0.1in}
\caption{The comparison of shape interpolation between SALD~\cite{atzmon_2021_sald} and our method on the SMAL dataset. (a) source shape. (b) target shape. (c) interpolation results of SALD~\cite{atzmon_2021_sald}. (d) our results. }
\label{fig:supp:shape_space:smal} 
\end{figure*}
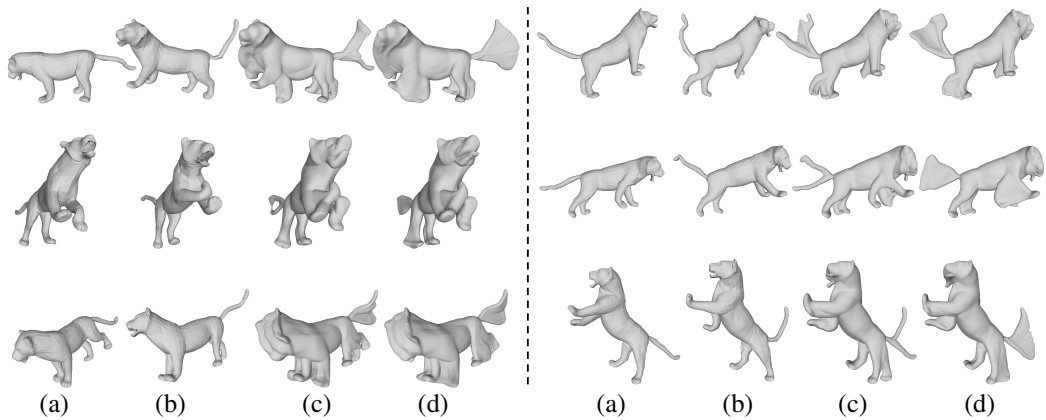

We show the shape interpolation results of the state-of-the-art implicit generator~\cite{atzmon_2021_sald} and our method in Figure~\ref{fig:supp:shape_space:dfaust} and Figure~\ref{fig:supp:shape_space:smal}. We show 30 interpolated shapes. By adding the proposed geometric deformation regularization loss and cycle-consistency regularization loss, our generator gives more meaningful interpolation results, which are important to the interpolation-guided registration.

\section{More Results of Shape Matching}
\label{Supp:Results:Corres}

\begin{figure*}
\centering
\begin{overpic}[width=1.0\textwidth]{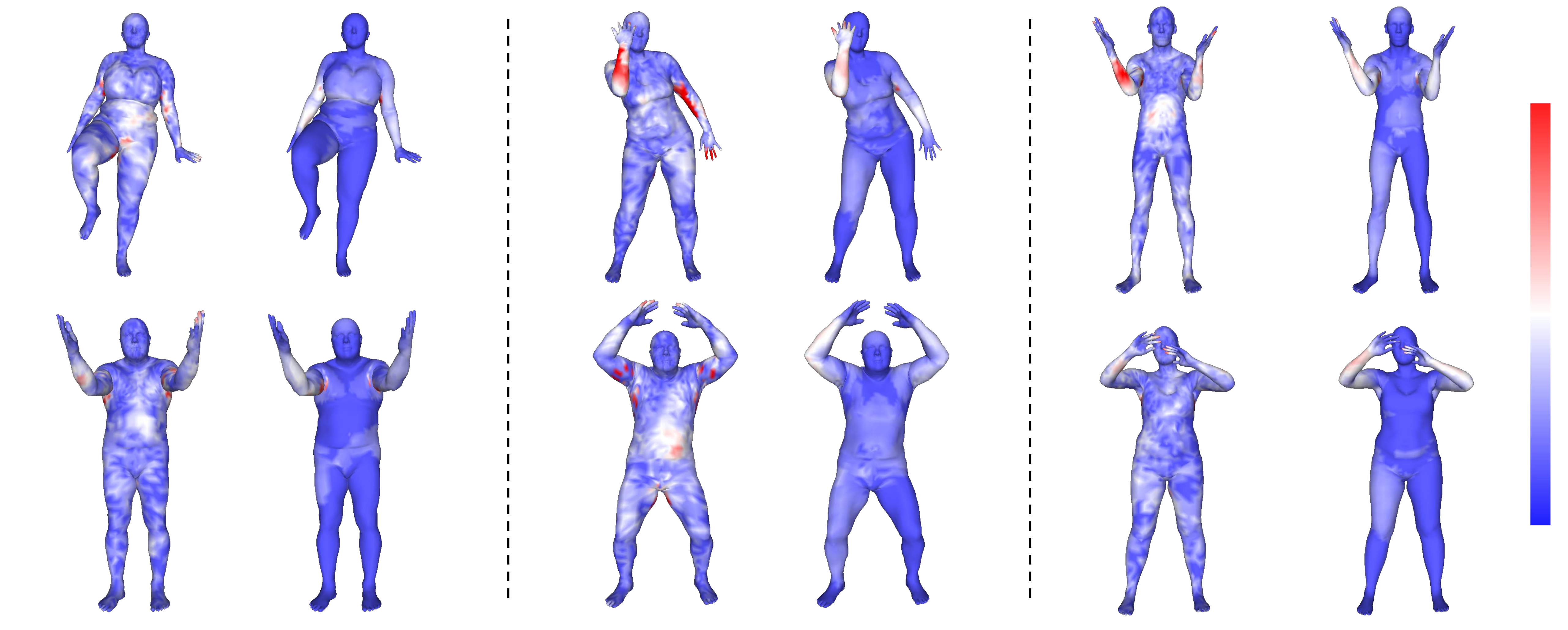}
\put(97.5, 4){\small{0}}
\put(96,  35){\small{0.15}}
\end{overpic}
\caption{The comparison of the correspondences between our method and NeuroMorph~\cite{Eisenberger_2021_neuromorph} on the DFAUST dataset. We show the correspondence errors on the target shapes. For each group of shapes, the left is from NeuroMorph, the right is our result. }
\label{fig:supp:corres:dfaust} 
\end{figure*}

\begin{figure*}
\centering
\begin{overpic}[width=1.\textwidth]{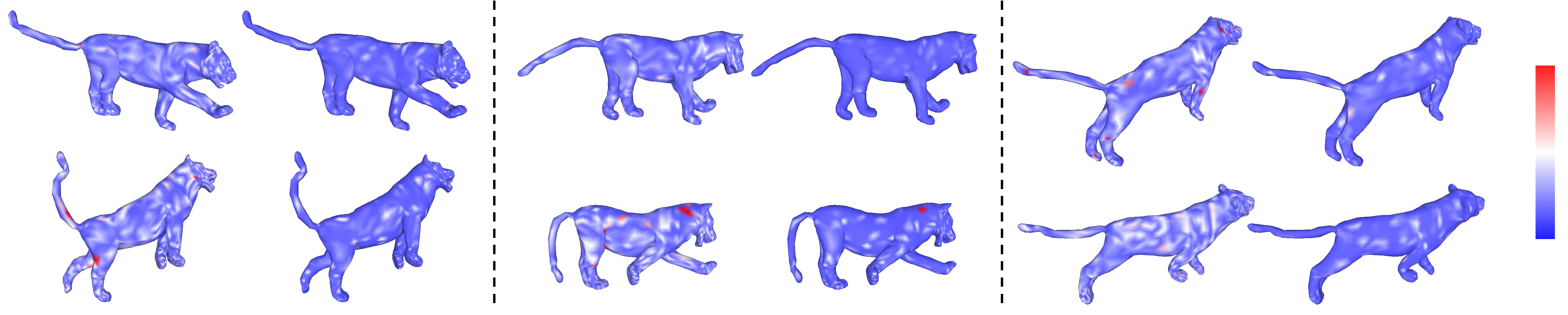}
\put(98, 2){\small{0}}
\put(96, 17){\small{0.15}}
\end{overpic}
\caption{The comparison of the correspondences between our method and NeuroMorph~\cite{Eisenberger_2021_neuromorph} on the SMAL dataset. We show the correspondence errors on the target shapes. For each group of shapes, the left is from NeuroMorph, the right is our result. }
\label{fig:supp:corres:smal} 
\end{figure*}

We show the correspondence results of NeuroMorph~\cite{Eisenberger_2021_neuromorph} and our method in Figure~\ref{fig:supp:corres:dfaust} and Figure~\ref{fig:supp:corres:smal}. Our method has lower errors compared to NeuroMorph.

\end{document}